\documentclass{article}

\usepackage[preprint]{corl_2026} % Use this for the initial submission.
\usepackage[utf8]{inputenc} % allow utf-8 input
\usepackage{mathtools} 

\usepackage{subcaption}
\usepackage[T1]{fontenc}    % use 8-bit T1 fonts
\usepackage{hyperref}       % hyperlinks
\usepackage{graphicx}
\usepackage{url}            % simple URL typesetting
\usepackage{booktabs}       % professional-quality tables
\usepackage{etoc}
\usepackage{amsfonts}   
\usepackage{amsmath}% blackboard math symbols
\usepackage{nicefrac}       % compact symbols for 1/2, etc.
\usepackage{microtype}      % microtypography
\usepackage{xcolor}         % colors
\usepackage{xcolor-material}
\usepackage{bbm} % for \mathbbm{1}

\usepackage[nameinlink,capitalise,noabbrev]{cleveref}
\crefname{appendix}{Appendix}{Appendices}
\Crefname{appendix}{Appendix}{Appendices}

\usepackage{algorithm}
\usepackage{algpseudocode}
\usepackage{float}
\usepackage{enumitem}
\definecolor{promptgray}{RGB}{230,230,230}    % Light gray background for prompts
\definecolor{responsegreen}{RGB}{220,255,220}   % Light green background for responses

\usepackage[most]{tcolorbox}
\tcbset{colback=gray!5, colframe=gray!50, fonttitle=\bfseries}
\usepackage{wrapfig}
\usepackage[table]{xcolor}
\usepackage{booktabs}
\usepackage{adjustbox}
\usepackage{multirow}
\usepackage{longtable}
\usepackage{pdflscape}
\usepackage{array}

\usepackage{makecell}

\newcolumntype{P}[1]{>{\raggedright\arraybackslash}p{#1}}

\tcbset{
  prompt/.style={
    colback=promptgray,
    colframe=promptgray,    % No solid boundary, same as background
    boxrule=0pt,
    arc=2mm,
    left=1pt,
    right=1pt,
    top=1pt,
    bottom=1pt,
    width=\linewidth,
    enhanced,
  },
  response/.style={
    colback=responsegreen,
    colframe=responsegreen, % No solid boundary, same as background
    boxrule=0pt,
    arc=2mm,
    left=1pt,
    right=1pt,
    top=1pt,
    bottom=1pt,
    width=\textwidth,
    enhanced,
  },
}

\definecolor{trendgreen}{RGB}{220,245,220}
\definecolor{driverpurple}{RGB}{235,225,250}
\definecolor{drivercyan}{RGB}{220,245,250}
\definecolor{sentred}{RGB}{250,225,225}
\definecolor{sentgreen}{RGB}{225,245,225}
\definecolor{horizonblue}{RGB}{225,235,250}
\definecolor{evidenceorange}{RGB}{255,235,210}
\definecolor{conflictyellow}{RGB}{255,245,200}
\definecolor{nonegray}{RGB}{235,235,235}

\newcommand{\baselinename}[1]{\textcolor{MaterialPurple900}{\textsf{#1}}}

\title{Coverage Aware Active Evaluation for Failure Discovery with Paired Systems}

\author{
Anjali Parashar$^{1}$\thanks{Corresponding author}, Rachel Luo$^{2}$, Apoorva Sharma$^{2}$, Sushant Veer$^{2}$, Edward Schmerling$^{2}$, \\ \textbf{Carson Sobolewski$^{1}$, Mingxin Yu$^{1}$, Chuchu Fan$^{1}$, Marco Pavone$^{2}$} \\
$^{1}$ Laboratory of Information \& Decision Systems (LIDS), MIT, USA \quad \quad $^{2}$ NVIDIA Research, USA\\
\texttt{anjalip,csobo,yumx35,chuchu@mit.edu} \\
\texttt{raluo, apoorvas, sveer, eschmerling, mpavone@nvidia.com} \\
}

\begin{document}
\maketitle

%===============================================================================

\begin{abstract}
%
%
% \sv{One can always find failures in sim, the trick is to find sim failures that are indicative of real-world failures. This point needs to be highlighted more in the abstract.}
%
% \rl{Agreed with Sushant's comments. I think it would be good to highlight the core challenge more clearly in the beginning (that sim failures are cheap to find, but only useful if they predict real-world failures). Maybe something like the below? }
%
Autonomous systems can fail in rare and heterogeneous ways, making real-world failure discovery difficult under limited testing budgets. Although cheaper proxies such as simulators, lower-fidelity systems, or related policies can be sampled extensively to find failures, proxy failures often do not transfer to the real world due to sim-to-real and system-to-system gaps. The key challenge is therefore to effectively leverage proxy system information for accurate prediction of severe target system failures. We propose an adaptive failure discovery method that combines proxy evaluations with limited target system results to guide scenario selection for target system testing. Our method learns a local predictor of target risk by correcting proxy failure signals using control-variate-inspired residual modeling. To find failures that are both likely and diverse, we combine this predictor with a support-aware mutual-information objective that favors realistic, well-supported regions while expanding coverage across failure modes. Across autonomous driving, manipulation, and quadruped velocity-tracking tasks, our method discovers up to 2x as many failures as random sampling and active-learning baselines, including severe and diverse failures missed by competing methods.

\end{abstract}

% Two or three meaningful keywords should be added here
\keywords{Testing \& evaluation, Adaptive Experimental Design} 

%===============================================================================

\section{Introduction}

Autonomous systems must be evaluated not only for average-case performance, but also for rare, severe, and heterogeneous failures that arise under specific scenario conditions. Such failures are especially important in safety-critical domains, where a small number of unanticipated edge cases can lead to unsafe behavior after deployment~\cite{dreossi2015efficient,esposito2005adaptive,corso2020scalable,corso2019adaptive,sinha2020neural}. However, discovering these failures through direct evaluation of the target system is difficult, since real-world or high-fidelity closed-loop tests can be expensive, time-consuming, and resource-constrained, and modern autonomous systems are often too complex to model analytically~\cite{parashar2024failure,sinha2025rate}. Cheap proxy systems, such as simulators, lower-fidelity platforms, related policies, or alternative evaluation processes, provide a natural way to scale failure search, but also introduce a central challenge over accuracy of failure estimation. Failures discovered in a proxy may reflect proxy-specific artifacts rather than true weaknesses of the target system. A scenario may appear critical because of simplified dynamics, fidelity mismatch, or distributional gaps, yet be benign on the target system; conversely, scenarios that appear safe in the proxy may still induce severe target-system failures. Thus, the goal is not merely to find failures cheaply, but to leverage proxy system carefully to accurately support  the prediction of predictive of target-system failures under a limited target-evaluation budget.

Existing approaches address only parts of this problem. Simulation-based testing can efficiently search low-cost models but may not transfer to the target system, while target-system active testing directly searches for critical scenarios but often does not exploit abundant proxy evaluations~\cite{dawson2023a,dawson2022robust,10669181,ren2023adaptsim,badithela2025reliable,sinha2025rate,parashar2024failure}. Methods that combine proxy and target evaluations can improve estimation of aggregate target-system metrics~\cite{luo2025leveraging,badithela2025reliable}, but these methods are often designed for Monte Carlo sampled data that can be ineffective for failure search. Failure discovery requires a different objective: actively selecting scenarios that reveal many distinct target-system failures. This also requires estimation of scenario specific metrics as opposed to aggregate metrics proposed by existing works.

\begin{figure}
    \centering
    \includegraphics[width=\linewidth]{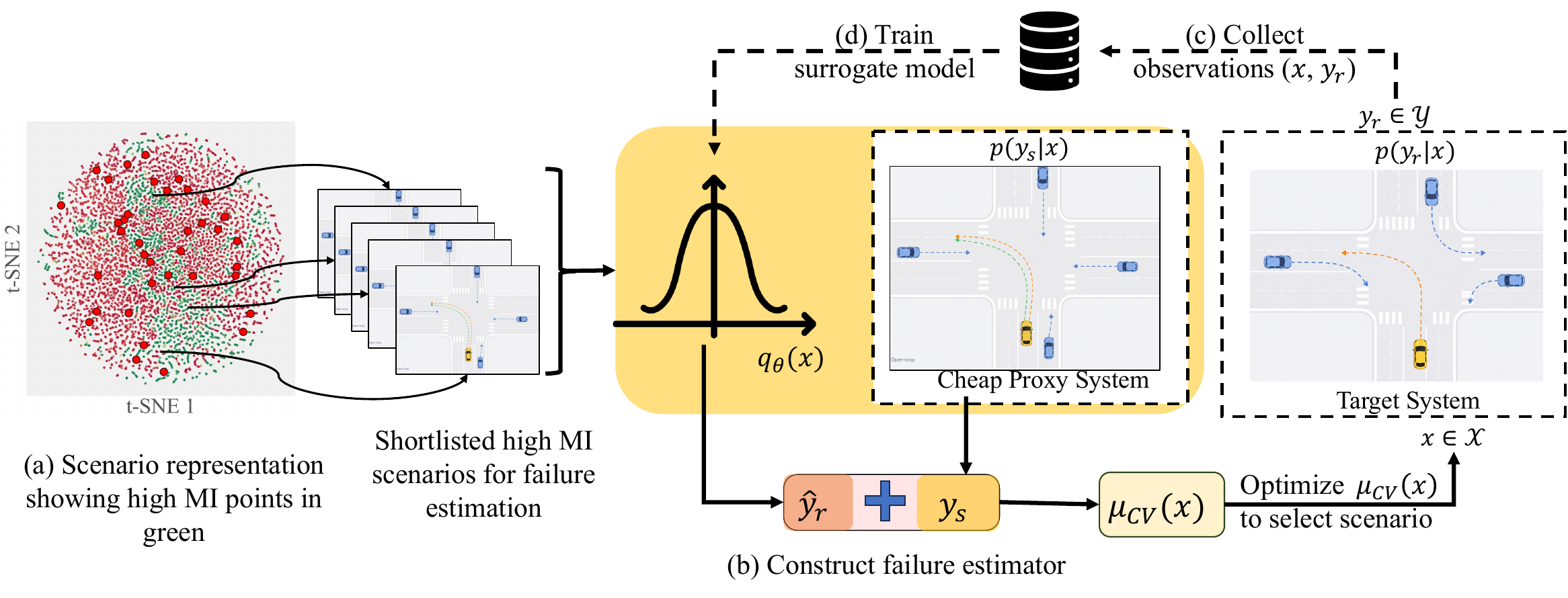}
    \caption{\textbf{Overview of approach.} Our method uses support-aware Mutual Information (\cref{eq:MI}) to subsample $K_{MI}$ scenarios $x$, which are evaluated on the cheap proxy system to get metrics $y_s$, and a surrogate model $q_{\theta}$ trained using data evaluated on target system. We subsequently optimize $\mu_{CV}(x)$ (\cref{eq:mu_cv}) over these sub sampled candidates to select the next batch of evaluation points as likely failure scenarios. The process is repeated until evaluation budget $B$ is exhausted.}
    \label{fig:process}
\end{figure}
In this work, we study adaptive failure discovery using cheap proxy evaluations and limited target-system tests. Our goal is to discover a large and diverse set of target-system failures under a fixed target-evaluation budget, where diversity corresponds to covering distinct regions of the scenario space, operating conditions, or failure modes. We propose an adaptive acquisition strategy that treats proxy evaluations as biased but informative signals. Rather than treating proxy failures as ground truth, our method learns a local failure predictor that estimates target-system risk by correcting proxy failure signals using limited target evaluations. This predictor is inspired by control variate estimation \citep{luo2025leveraging}, where proxy observations provide a low-cost signal correlated with target behavior, while target evaluations correct for local proxy-target mismatch. The resulting estimate prioritizes scenarios that are likely to fail on the target system, rather than merely fail in the proxy. Our contributions are as follows:
\begin{enumerate}[leftmargin=1em,itemsep=0pt]
    \item We formulate proxy-guided failure discovery as an adaptive scenario selection problem, where the goal is to use limited target-system evaluations to discover many distinct target failure modes.
    \item We propose a local, proxy-corrected failure predictor that uses limited paired proxy-target evaluations to identify scenarios likely to fail on the target system, rather than merely in the proxy evaluator.
    \item We introduce a support-aware mutual-information acquisition strategy that balances target-failure likelihood with coverage of realistic and underexplored regions of the scenario space, encouraging acquisition of novel failure scenarios.
    \item We evaluate our method across autonomous driving, multi-fidelity manipulation, and quadruped velocity-tracking tasks, showing that it discovers more target-system failures and reveals a broader range of critical failure modes than random sampling and active-learning baselines.
\end{enumerate}

Across these domains, our method discovers upto twice as many target-system failures than the strongest baseline under the same target-evaluation budget, discovering several low and high severity failures, where the baselines completely miss discovery of high severity failures. Our method also achieves highest diversity across all tasks and baselines, under a pairwise-distance-based coverage metric. These results show that cheap proxy evaluations can substantially improve failure discovery when used to guide target-system testing in an adaptive manner, rather than as direct substitutes for target-system evaluation.

\section{Related Work}

\paragraph{Simulation-based failure discovery.}
Simulation enables scalable, low-cost testing of autonomous systems via adaptive sampling, stress testing, and failure discovery~\cite{dawson2023a,dawson2022robust}. However, these methods are sensitive to sim-to-real gaps and model misspecification, so scenarios appearing safe in simulation may still fail on the real system~\cite{10669181,ren2023adaptsim,parashar2024failure}. We address this by treating proxy evaluations as biased but informative signals, and using limited target evaluations to adaptively correct the proxy-to-target relationship during failure discovery.

\paragraph{Sample-efficient target-system testing.}
A complementary line of work focuses on failure discovery directly on the target system using expert-designed scenario sets, or surrogate based Bayesian optimization, or active learning under limited budgets~\cite{parashar2024failure,parashar2025cost, anwar2025efficient}. Surrogate based approaches can struggle to scale when scenarios are high-dimensional or failure landscapes are complex~\cite{sinha2025rate,parashar2025cost}. Our method differs by not depending on restrictive model choices while modeling target-system risk.

\paragraph{Combining proxy and target evaluations.}
Recent works use Control variates and related variance-reduction methods to use correlated proxy signals for improved estimation of aggregate real-world metrics under limited target evaluations~\cite{luo2025leveraging,badithela2025reliable}. These methods are typically designed for Monte Carlo estimation of global quantities, which is inefficient for discovering rare failures. Our goal instead is to actively select scenarios that reveal  many target-system failures, motivating an adaptive acquisition strategy that uses proxy-target correction for both failure estimation and scenario selection.

\paragraph{Bayesian experimental design and information-based acquisition.}
Bayesian experimental design approaches select informative evaluations under limited budgets~\cite{rainforth2024modern,Chaloner1995BayesianReview}, often via Mutual Information (MI)~\cite{cao2021bayesian,parashar2026seed}. However, for failure discovery, uncertainty alone may target non-critical scenarios, while exploitation alone may produce duplicate failures. We therefore combine a support-aware mutual-information objective for exploring under-covered regions with a proxy-corrected risk estimate to prioritize severe failures.

\paragraph{Positioning of this work.}
Our work connects these lines by combining cheap proxy evaluations, limited target-system tests, and information-based acquisition. Unlike simulation-only testing, we do not assume proxy failures transfer directly to the target. Unlike target-only active testing, we leverage proxy evaluations to guide search. Unlike variance-reduced estimation, our goal is not to estimate an aggregate metric but to select evaluations that reveal diverse failure modes. Together, these components enable sample-efficient discovery of distinct target-system failures.

% Our work connects these lines of research by combining cheap proxy evaluations, limited target-system tests, and information-based acquisition for adaptive failure discovery. Unlike simulation-only testing, we do not assume proxy failures transfer directly to the target. Unlike target-only active testing, we use available proxy evaluations to support scenario search. Unlike variance-reduced performance estimation, our goal is not only to estimate an aggregate metric, but to select target evaluations that reveal diverse failure modes. Together, these components enable sample-efficient discovery of distinct target-system failures under limited testing budgets.

\section{Problem Statement}

We consider a black-box dynamical system that maps scenario parameters $x \in \mathcal{X}$ to trajectory rollouts. Our goal is to discover failures of this \textit{target} system, which can be expensive to evaluate. Failure is  measured by a safety metric $y_r \in \mathcal{Y}$, where,  without loss of generality, larger values indicate less safe behavior. We assume access to a cheaper \textit{proxy} system that provides low-cost observations $y_s$ for the same scenarios. Rollout stochasticity or environmental variations induce distributions $p(y_s \mid x)$ and $p(y_r \mid x)$ for these systems. Because proxy evaluations are inexpensive, $p(y_s \mid x)$ can be estimated by sampling, while estimating $p(y_r \mid x)$ remains difficult due to high cost of evaluations (\cref{fig:process}).

Our objective is to use paired proxy and target information to discover a large number of target system failures, spanning diverse scenarios or operating conditions, under a target evaluation budget $B$. For scenario $x$, let $\mu_r(x) \coloneqq \mathbb{E}_{p(y_r \mid x)}[y_r]$. We define failure as $\mu_r(x) \geq \gamma$, where $\gamma$ is a user-specified severity threshold. The goal is to identify a diverse set of failures
$
X_B = \{x^i \in \mathcal{X} \mid \mu_r(x^i) \geq \gamma\}_{i=1}^K.
$
We promote diversity using mutual information (MI), which is commonly used to quantify epistemic uncertainty from limited evaluations~\citep{parashar2026seed,cao2021bayesian,rainforth2024modern,Chaloner1995BayesianReview}.

To maintain the evaluation budget, we formulate failure discovery as adaptive scenario sampling using Bayesian Experimental Design (BED)~\cite{rainforth2024modern}. At iteration $k \leq B$, given observed target evaluations $\mathcal{D}_k={(x^i,y_r^i)}_{i=1}^k$, we select the next scenario by maximizing an acquisition function $\alpha:\mathcal{X}\to\mathbb{R}$
as $ 
x_{k+1} = \arg\max_{x\in\mathcal{X}} \alpha(x).
$
Our acquisition must balance two goals: selecting scenarios likely to satisfy $\mu_r(x)\geq\gamma$, and covering diverse operating conditions through high MI. We next describe the design of this acquisition strategy.

\section{Data acquisition strategy for meeting dual evaluation objectives}
Because the target system is black-box, directly estimating $\mu_r$ is difficult. We can instead learn a surrogate,  $q_\theta(x)$ from observed target evaluations $\mathcal{D}_k=\{(x^i,y_r^i)\}_{i=1}^k$ to approximate $p(y_r|x)$, and use it to guide future scenario selection. However, target-only surrogates ignore the readily available proxy data. To combine predictions from $q_\theta(x)$ learnt using limited target evaluations with biased, noisy proxy signals for accurate failure prediction, we introduce a local control-variate predictor $\mu_{\mathrm{CV}}$ in \cref{sec:cv}.
To promote scenario diversity, we introduce a support-aware MI term $I(Z_x;R_x\mid x,\mathcal{D}_t)$ in \cref{sec:mi}, where $Z_x$ captures the evaluated data support and $R_x$ denotes the potential support improvement from candidate $x$. Combining this diversity objective with the control-variate failure predictor gives the high level objective for data acquisition:
% \begin{equation}\label{eq:alpha}
% \alpha(x\mid\mathcal{D}_{t}) =\mu_{\mathrm{CV}}(x) + I(Z;R_x\mid x,\mathcal{D}_t).
% \end{equation}
\begin{equation}\label{eq:alpha}
\alpha(x\mid\mathcal{D}_{t}) =
\underbrace{\mu_{\mathrm{CV}}(x)}_{\text{failure estimation}}
+
\underbrace{I(Z_x;R_x\mid x,\mathcal{D}_t)}_{\text{scenario diversity}}.
\end{equation}
The two objectives cumulatively represent candidates that are both likely target failures and informative for expanding support. We define $\mu_{\mathrm{CV}}$ and the support-aware MI term next. 

\subsection{Combining target and proxy system information for failure estimation}\label{sec:cv}
We first describe the construction of $\mu_{CV}$ for failure estimation. Consider a candidate scenario $x \in \mathcal{X}$ for target-system testing, where the goal is to identify scenarios with $\mu_r(x) \geq \gamma$. As discussed above, relying only on the surrogate $q_{\theta}$ can be inefficient because it is learned from limited target data and ignores proxy evaluations $p(y_s)$. We therefore propose a local, scenario-specific mean estimator $\mu_{\text{CV}}(x)$ based on the Control Variate (CV) method~\citep{luo2025leveraging}. 
By incorporating $\mu_{CV}$ in $\alpha$, we enable a more accurate failure estimation at a given $x$, than can be achieved using just $Y_r$ with limited samples, or, using a large number of proxy samples $Y_s$.
This estimator uses proxy evaluations $p(y_s)$ as additional information to improve the target mean estimate obtained from $q_{\theta}$. For a candidate scenario $x$, we construct a neighborhood $\mathcal{B}_x$ and sample $N$ scenarios $x^i \sim \mathcal{B}_x$ for surrogate prediction. Here, $\mathcal{B}_x$ can denote a categorical scenario type or a Euclidean ball in continuous scenario space, as detailed in \cref{app:local-region}. Within this region, we uniformly sample at most $N =n+k$ datapoints: paired evaluations $(x^i,\hat{y}_r^i,y_s^i)_{i=1}^n$, where $\hat{y}_r \sim q_{\theta}(x)$, and additional proxy evaluations $(x^j,y_s^j)_{j=1}^k$. Using this data, we approximate $\mu_r(x)$ using $\mu_{CV}(x)$, a local mean estimate of the metric at $x$, as:
% \begin{equation}\label{eq:mu_cv}
% \begin{aligned}
%      \mu_{CV}(x) = \frac{1}{n} \sum_{i=1}^{n} (\hat{y}_r^i-\boldsymbol{\beta} y_s^i )+ \frac{1}{k} \sum_{j=1}^{k} \boldsymbol{\beta} y_s^i, \\
%      \boldsymbol{\beta} = \bigg( \frac{k}{k+n}\bigg) \text{Var}(Y_s)^{-1} \text{Cov}(Y_r,Y_s),
% \end{aligned}
% \end{equation}
\begin{equation}\label{eq:mu_cv}
\mu_{CV}(x) = \frac{1}{n} \sum_{i=1}^{n} (\hat{y}_r^i-\boldsymbol{\beta} y_s^i )
+ \frac{1}{k} \sum_{j=1}^{k} \boldsymbol{\beta} y_s^i,
~~ \text{where} ~~
\boldsymbol{\beta} = \bigg( \frac{k}{k+n}\bigg) \text{Var}(Y_s)^{-1} \text{Cov}(Y_r,Y_s).
\end{equation}
Here $Y_s = (y_s^j)_{j=1}^{n+k}, Y_r = (\hat{y}_r^j)_{j=1}^{n}$, and $\beta$  adjusts the mean estimated using proxy metrics based on the correlation between target and proxy distributions $q_{\theta}(x)$ and $p(y_s|x)$ for $x \in \mathcal{B}_x$. This results in estimator $\mu_{CV}$ that corresponds to  minimum variance estimator of the failure statistics in the local region (see \cref{app:cv-global} for details). \cref{fig:cv-radius} shows the application of this approach for failure prediction on a 1D toy task. 

\subsection{Formalizing scenario diversity for exploration}\label{sec:mi}
To encourage scenario diversity in adaptive data acquisition, we aim to prioritize scenarios away from observed scenarios. To this end, we construct a support aware MI, such that maximizing MI identifies novel scenarios. As a first step, we explicitly identify scenario space covered using $\mathcal{D}_t$. This can be done by clustering observed data in $K$ clusters in the scenario space. For a an unseen potential candidate $x$, let $Z_x \in [1,\dots,K, \text{new}]$ denote cluster assignment at $x$, where $Z_x =  \text{new}$ implies formation of new cluster by evaluating at $x$, implying that the point is not within the covered scenario space. Tied to $x$, we also define a support discovery variable $R_x \in \{0,1\}$, where $R_x=1$ denotes that $x$ reveals new support. Thus, our goal of encouraging diversity in scenario can be defined as discovery of novel candidates far from support and those within support with uncertainty on cluster assignment. Mathematically, this can be naturally captured by MI as $I(Z_x;R_x|\mathcal{D}_t,x)$ in a principled manner, defined using entropy $H$ as:
\begin{equation}\label{eq:MI}
    I(Z_x;R_x \mid D_t, x)
=
H(Z_x|\mathcal{D}_t,x) - H(Z_x|R_x,\mathcal{D}_t,x).
\end{equation}

To enable a direct and tractable connection between $R_x$ and $Z_x$, we model the distribution over $R_x$ as $p(R_x|x,\mathcal{D}_t) = \sum_{k=1}^Kp(R_x|Z_x=k) p(Z_x=k|x,\mathcal{D}_t)$. Here, prior $p(Z_x=k|\mathcal{D}_t,x)$ controls probability of assignment of a point $x$ to cluster $k$, and we define the likelihood  $p(R_x=1|Z_x=k)$ to be very low for scenarios well within one of the $k$ clusters, and set it to be high for $Z_x = \text{new}$ (see \cref{app:MI-hyperparam} for details). Under these modeling choices, the MI term in \cref{eq:MI} prioritizes scenarios for which cluster assignment is ambiguous, due to lying between cluster boundaries, and scenarios far from existing support. Thus, evaluating on these points enables meaningful support expansion, meeting our goal of scenario diversity.

% \cref{fig:process} visualize the MI for a collection of randomly sampled  points for a toy 2D example, showing high MI for unobserved data far from evaluated data shown in red. 

Unlike GP-based approaches such as \cite{sinha2025rate}, our method scales to high-dimensional scenarios because it is agnostic to the surrogate architecture $q_{\theta}$ and encodes exploration directly through support. This enables scaling beyond evaluation budgets of only tens of samples. In \cref{fig:nuplan-MI}, we illustrate this behavior on a large-scale, high-dimensional dataset, where the method prioritizes unexplored regions under different supports induced by evaluated samples $\mathcal{D}_t$.

\subsection{Batched data acquisition for diverse failure discovery}\label{sec:acf}
% Using the two main components, we propose a complete acquisition function $\alpha$ given by:
% \begin{equation}\label{eq:alpha}
%     \alpha(x|\mathcal{D}_{t}) = \mu_{CV}(x) + \gamma JS (M;y|x,\mathcal{D}_t) + \beta I(\theta_r;y|x,\mathcal{D}_t),
% \end{equation}

% \begin{equation}\label{eq:alpha}
%     \alpha(x|\mathcal{D}_{t}) = \mu_{CV}(x) +I(Z;R_x|x,\mathcal{D}_t),
% \end{equation}
% where $\mathcal{D}_t = (x^i,y_r^i)_{i=1}^t$ denotes a sequentially updated dataset for $t\leq B$, and $y_r^i$ corresponds to target system evaluation on scenario $x^i$. For a single sample evaluation, at each iteration we select sample $x^i = \arg\max_{x \in \mathcal{X}} \alpha(x)\}_{i=1}^B$. 
Using the two main components in \cref{eq:alpha}, we can optimize $\alpha$ for scenario discovery in a continuous domain. However, in our experimental validation, the scenario space $\mathcal{X}$ is a collection of discrete samples, and our objective is to get $\mu_{CV} \geq \gamma$. Hence, we transform the optimization procedure into a severity adjusted data sampling rule with batched data acquisition of $b$ scenarios at each iteration, for a budget of $B = Tb$ samples. The batch size $b$ is assumed to be user defined, and selected to be large enough to provide sufficient dataset for updating $q_{\theta}$ based on scenario dimension, and low enough that we do not surpass the budget $B$. At each step $t$, we select top-$K_{MI}$ candidates that maximize $I(Z_x;R_x|x,\mathcal{D}_t)$, followed by construction of the set $\Omega_t = \{x|\mu_{CV}(x) \geq \gamma\}$. We select $b$ candidates from $\Omega_t$ by clustering the samples into $b$ clusters and selecting candidate closest to cluster center for each, to further reinforce diversity within the batch. We note that the final algorithm is not a complete equivalent of \cref{eq:alpha}, but represents a constrained batched optimization equivalent instead, with the said constraints and batching procedures enabled for efficiency. \cref{algo:algo-AL} summarizes the data acquisition procedure.
% and \cref{fig:toy-2d} shows an application of our approach on a toy 2D task with two distinct failure modes, where proxy system under represents one of the failure modes. 

% The term $\beta$ refers to \textit{degree of curiosity} and controls the degree of epistemic exploration. We use the paired data evaluations to calibrate $\beta$ to ensure equal representation of exploration and exploitation.

% Appendix~\ref{app:gap-energy} provides probabilistic and geometric perspective on the discrepancies captured using correlation ($\rho$), and $I(M,y|x,\mathcal{D})$, and why we need both. 

\begin{algorithm}[H]
\caption{Adaptive scenario discovery using \eqref{eq:alpha}.}
\label{algo:algo-AL}
\begin{algorithmic}[1]
\State \textbf{Input:} Scenario space $\mathcal{X}$, initial paired $\mathcal{D}_0 = \{x^i,y^r_i\}_{i=1}^{N_0}$ 
% and MC sampled unpaired dataset $\mathcal{D}_u$ (optional). 
\State Initialize surrogate models $q(x)$ trained using $\mathcal{D}_0$.
\For{$t = 0$ to $T-1$} 
    \State Select scenario $\mathcal{X}_u = \{x^i| \arg \max_{x^i\in \mathcal{X}} I(Z;R_x|x^i,\mathcal{D}_t)\}_{i=1}^{K_{MI}}$ for $I$ in \cref{eq:MI},
    \State Construct $\Omega_t=\{x\mid \mu_{CV}(x) \geq \gamma\}$ for $\mu_{CV}$ in \cref{eq:mu_cv}. 
    \State Select $(x^i)_{i=1}^b$ for $x^i \in \Omega_t$, closest to cluster centers for $b$ cluster on $\Omega_t$. 
    \State Collect $(y_r^i)_{i=1}^b$ for $(x^i)_{i=1}^b$  using target system evaluation
    \State Update dataset: $\mathcal{D}_{t+1} = \mathcal{D}_{t} \cup \{(x^{i}, y^{i}_r )\}_{i=1}^b$
    \State Update surrogate models $q_{\theta}$, using $\mathcal{D}_{t+1}$.
\EndFor
\end{algorithmic}
\end{algorithm}

\section{Experimental validation}
The central hypotheses of this paper are: (1) our approach can identify a concise set of scenarios that reveal failures across a range of severity levels and diverse operating conditions, and (2) it scales to larger evaluation budgets for high-dimensional and complex scenarios. We validate these hypotheses on four autonomous system tasks: \textsf{nuPlan}, \textsf{SIMPLER}, \textsf{Quadruped},  and \textsf{KITTI}, spanning autonomous driving, manipulation, and quadruped velocity tracking. These tasks represent a broad range of scenario specifications, proxy-target correlation structures, and evaluation costs.

\paragraph{Tasks.} In \textsf{nuPlan}, we evaluate motion-planning behavior using driving scenarios from the \textsf{nuPlan} dataset~\citep{karnchanachari2024towards}, with open-loop playback as the proxy and closed-loop rollouts as the target. Scenarios are encoded in a
384-dimensional space, and we use Time-to-Collision (TTC) as the failure metric. In \textsf{SIMPLER}, we evaluate the language-conditioned manipulation policy \textsf{RT-1x}~\citep{open_x_embodiment_rt_x_2023} on a grasp and lift task~\citep{li24simpler}, where proxy and target correspond to two photorealistic simulator environments differing in visual specifications; scenarios are parameterized by a
19-dimensional space encoding object positions, distractors, and camera pose. In \textsf{Quadruped}, we evaluate a \textsf{Unitree Go2} quadruped~\citep{unitree_go2pro} on a 3 dimensional command velocity tracking task, using a \textsf{MuJoCo}-based simulation with an RL policy~\citep{unitree_rl_mjlab} as the proxy, demonstrating real-world applicability of our approach. In \textsf{KITTI}, we analyze perception failures in autonomous driving,  using a continuous failure score from GT vehicle coverage on a real world driving dataset KITTI \cite{geiger2012kitti}, paired with its synthetic clone, Virtual KITTI \cite{cabon2020virtual}. These results demonstrate that our method utilizes proxy information effectively, and improves failure diversity. We also provide a supplementary video showing our approach for failure discovery on \textsf{Quadruped} task. 

We use BNN for modeling $q_{\theta}$ for \textsf{nuPlan, SIMPLER, KITTI} and MLP for \textsf{Quadruped} to show that unlike baselines such as \cite{sinha2025rate,parashar2025cost}, our method is not sensitive to the choice of surrogate model. \cref{app:experiment-details} provides additional implementation details for all tasks, and \cref{app:scenario} shows visualization of exemplar failure scenarios discovered by our approach, in addition to dominant failure modes observed.

\paragraph{Proxy system budget.} We control proxy budget using a fixed proxy-compute budget $B_p=N_p c_p$ across tasks, adjusting $N_p$ inversely with per-scenario proxy runtime $c_p$. Total measurable computational cost is $C=N_t c_t+N_p c_p+C_{\mathrm{train}}^t+C_{\mathrm{train}}^p,$ where $C_{\mathrm{train}}^p$ (surrogate training for proxy) is optionally used when proxy data is scarce, and $C_{\mathrm{train}}^p,C_{\mathrm{train}}^t$ are much smaller than evaluation costs, therefore $C \approx N_t c_t+N_p c_p$. We consider tasks with a variety of proxy runtimes, for example, SIMPLER has high $c_p$, $\approx$ 83s/eval vs.\ 2s/eval for Quadruped. This limits proxy data in SIMPLER to a total 110 samples (50 initial and 60 sequential) whereas in Quadruped, we are able to use 2,000 samples.  In practice, target testing also incurs resource, and potential-damage costs that cannot be reduced to a scalar cost. These task-specific constraints motivate conservative target testing budget $N_t = N_0 + B$, reflecting our focus on fixed target-evaluation budgets. 
\paragraph{Baselines.} We compare our approach against the following baselines: \baselinename{Random} (na\"ive MC sampling); \baselinename{BAMS}~\citep{sinha2025rate}, which uses a GP surrogate with multi-fidelity variance for acquisition; \baselinename{BNN-C/GP-C}~\citep{parashar2024failure}, which acquires worst-case scenarios via clustering with a BNN or GP surrogate; and \baselinename{BNN-CV}, an ablation of our method without MI-driven exploration. We use GP for \textsf{SIMPLER}, and BNN for \textsf{nuPlan} for \baselinename{BNN-C/GP-C}.

% In \textsf{nuPlan} task, we additionally consider \baselinename{BNN-Random, BNN-IS}, which correspond to BNN surrogate trained using random sampling and importance sampling respectively, to test the prediction accuracy of surrogate learnt using different data acquisition strategies.

\paragraph{Metrics.}
To validate our first hypothesis, we evaluate all methods using three metrics. \textit{Cumulative average metric} measures the average severity of discovered scenarios up to budget $k$, computed as $\frac{1}{k}\sum_{k=1}^{B} y_r^k$. \textit{Cumulative coverage} measures diversity among high-failure scenarios $\mathcal{F}_{\mathrm{true}} = \{x \in X_B : y_r \geq \gamma\}$ via their average pairwise distance,
$
C_{k}(\mathcal{F}_{\mathrm{true}}) = \frac{1}{k} \sum_{i,j \in [1,\dots,k]} \left\| x^i - x^j \right\|^2_2$ for $1 \leq k \leq B,
$
where larger values indicate failures spread across the scenario space rather than collapsing onto a single region. \textit{Positive samples} $P = \lvert \mathcal{F}_{\mathrm{true}} \rvert$ counts discovered failures at varying severity thresholds $\gamma$, reported for \textsf{nuPlan} in \cref{tab:ps-tasks}.

% \paragraph{Metrics.} To validate our central hypothesis, we construct three metrics, namely \textit{Cumulative average metric},  \textit{Cumulative coverage} and \textit{Positive samples} that we test across baselines on \textsf{nuPlan} and \textsf{SIMPLER} tasks.  
% \textit{Cumulative average metric} measures optimality of scenarios discovered so far using $\frac{1}{k} \sum_{k=1}^B y_r^k$. 
% For set of scenarios $\mathcal{D}_B$ selected by a given method, we compute the average pairwise distance among retained high-failure scenarios in a set $\mathcal{F}_{\mathrm{true}} = \{x \in X_B: y_r \geq \gamma\}$.  Given a pair of scenarios  $x^i,x^j \in \mathcal{F}_{\mathrm{true}}$, we define cumulative coverage as
% $C_{k}(\mathcal{F}_{\mathrm{true}})
% =
% \frac{1}{k}
% \sum_{i,j\in [1,\dots,k]}
% \left\|
% x^i-x^j
% \right\|^2_2, $
% for $1 \leq k \leq B$. 
% A larger value indicates that the observed failures are more spread out in the scenario space, suggesting that the method discovers diverse failure modes rather than collapsing onto a single local region. We also report \textit{positive samples} $P = \lvert \mathcal{F}_{\mathrm{true}}\rvert$ for \textsf{nuPlan} at various values of $\gamma$ to interpret diversity of failures at various severity levels, shown in \cref{tab:ps-tasks}. 

\paragraph{Ablations} We perform following ablations on our approach using the \textsf{nuPlan} task, (1) varying size of initial dataset (\cref{app:nuplan-ablations}) and form of sampling to demonstrate the utility of support aware MI (\cref{fig:nuplan-MI}), (2) ablations on formulation of $\mu_{CV}$ in \cref{eq:mu_cv}  (\cref{app:nuplan-ablations}).  On \textsf{KITTI}, we additionally demonstrate component wise ablation of our approach, isolating surrogate, proxy contribution, MI and clustering (Table~\ref{tab:kitti_component_ablation}). MI-only (MI) yields high coverage but low failure discovery; no clustering (Greedy) reduces coverage due to lesser intra-batch diversity and suffers from mode collapse; no MI with clustering ($-$MI-C) under-performs on high-severity failure discovery and coverage; and no MI and no clustering ($-$MI-G) further reduces coverage. Thus, ablations show that MI promotes support coverage, CV improves failure targeting, and clustering preserves within-batch diversity.

\begin{table}[t]
\centering
\caption{Positive Samples for different severity levels for \textsf{nuPlan} (mean $\pm$ std across four seeds).}
\label{tab:ps-tasks}
\resizebox{\linewidth}{!}{
\begin{tabular}{llccccc}
\toprule
Task & $\gamma$ & Random & BNN-C & BAMS & BNN-CV & Our \\
\midrule
\multirow{3}{*}{nuPlan }
& 0.3 & 0.23 $\pm$ 0.01 & 0.25 $\pm$ 0.01 & 0.28 $\pm$ 0.01 & 0.27 $\pm$ 0.01  & \textbf{0.33 $\pm$ 0.01}  \\
& 0.2 & 0.22 $\pm$ 0.00 & 0.24 $\pm$ 0.02  & 0.27 $\pm$ 0.01  & 0.26 $\pm$ 0.01  & \textbf{0.32 $\pm$ 0.01}  \\
& 0.05 & 0.21 $\pm$ 0.01  & 0.21 $\pm$ 0.02  & 0.25 $\pm$ 0.01  & 0.22 $\pm$ 0.02  & \textbf{0.29 $\pm$ 0.01} \\
\midrule
% \multirow{3}{*}{SIMPLER}
% & 0.5 & -- & -- & -- & -- & --  \\
% & 0.3 & -- & -- & -- & -- & --  \\
% & 0.1 & -- & -- & -- & -- & --  \\
% \bottomrule
\end{tabular}
}
\end{table}

\subsection{\textsf{nuPlan}}\label{sec:nuplan}

\begin{figure}
    \centering
    \includegraphics[width=0.9\linewidth]{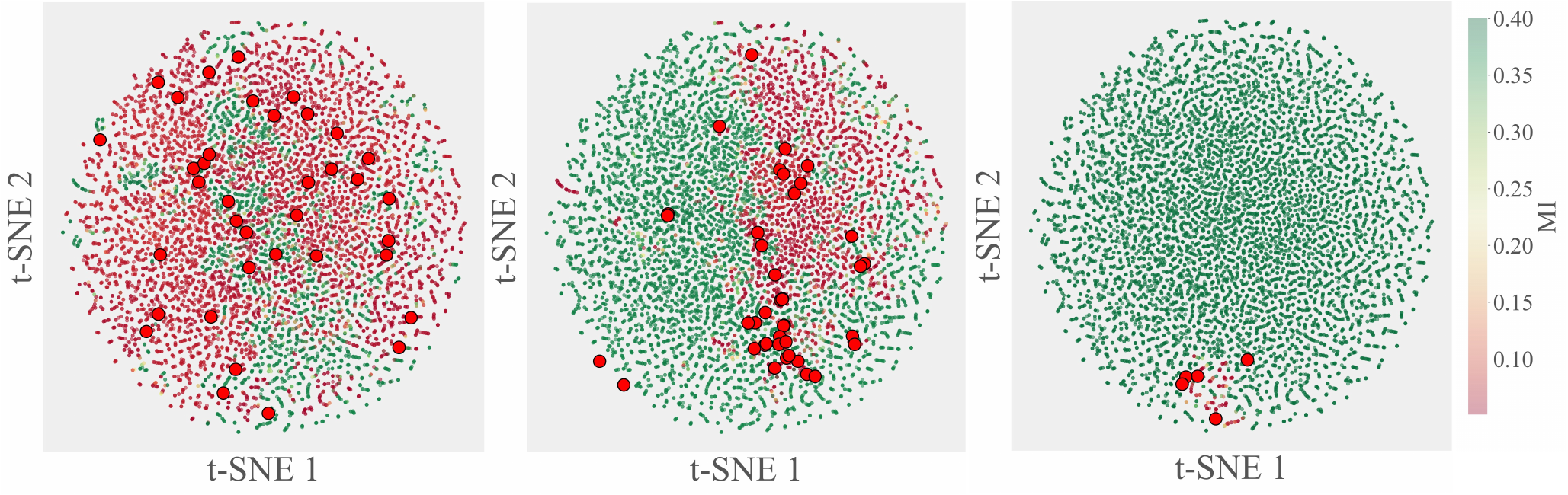}
    \caption{\textbf{MI driven exploration for different forms of data acquisition on \textsf{nuPlan} task}.  2D T-SNE embeddings of scenario dataset showing high and low MI regions  for 200 datapoints sampled from (1) small concentrated region of low failures, (2) MC sampling, (3) data recovered from a specific type of scenarios. The plots show that our support aware MI strategy successfully identifies unobserved regions agnostic to the choice of sampling strategy.}
    \label{fig:nuplan-MI}
\end{figure}

\begin{figure}
    \centering
    \includegraphics[width=0.9\linewidth]{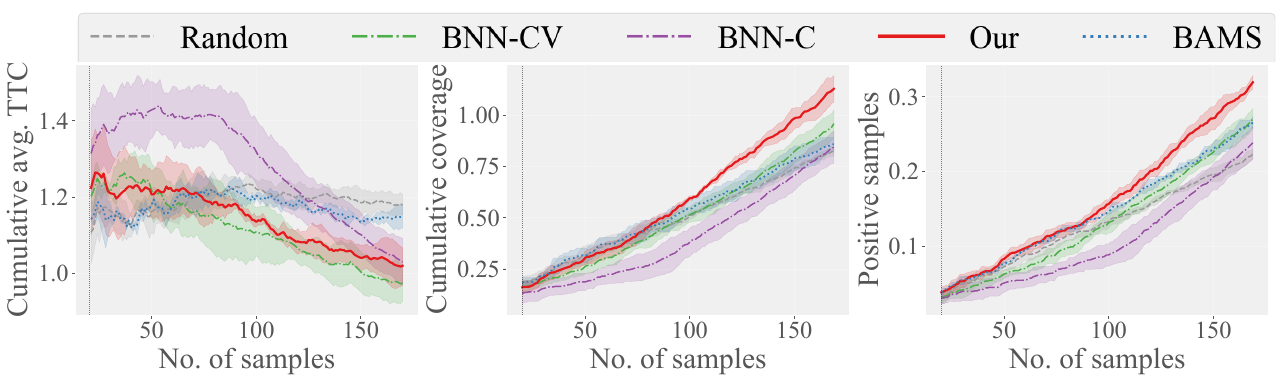}
    \caption{\textbf{Baseline comparison} for \textsf{nuPlan} task, (avg $\pm$ std across four seeds) for $B=170$ samples, with $20$ initial random samples for training surrogate, and 150  samples acquired in batches of $b=5$.}
    \label{fig:nuplan-metrics}
\end{figure}

\cref{fig:nuplan-metrics} shows that our approach achieves lower cumulative average metric than all baselines except \baselinename{BNN-CV}, which repeatedly samples the same failures, hurting both diversity and failure count. Our approach outperforms all baselines on both cumulative coverage and number of failures discovered, where $\mathcal{F}$ is defined by closed-loop TTC $< 0.3$. In practice, we often lack control over how the initial scenario set is collected. \cref{fig:nuplan-MI} visualizes MI values across the full \textsf{nuPlan} scenario embedding space, showing that our support-aware MI consistently prioritizes unexplored regions regardless of how prior samples were collected, providing a principled measure of scenario diversity.

\subsection{\textsf{Quadruped} }\label{sec:quadruped}
\begin{figure}
    \centering
    \includegraphics[width=0.9\linewidth]{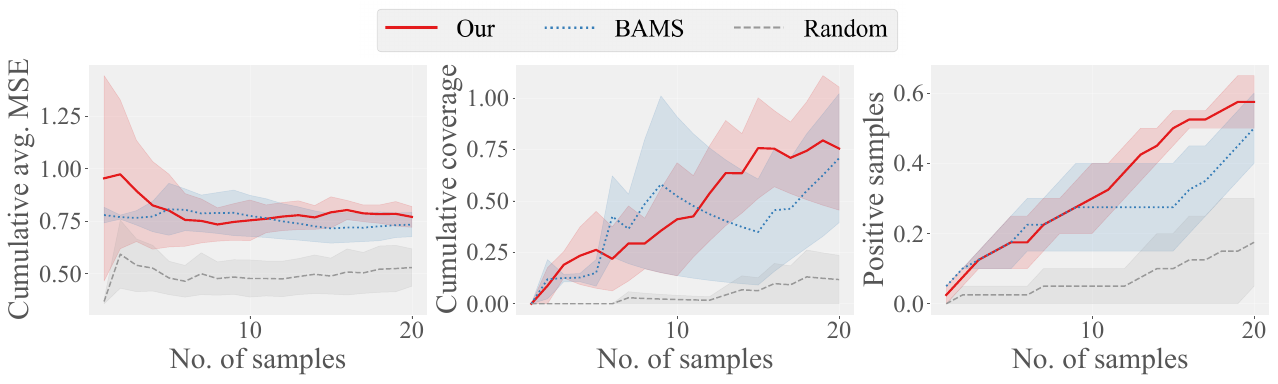}
    \caption{\textbf{Baseline comparison} for  \textsf{Quadruped} task, (avg $\pm$ std, two seeds) for $B=30$ samples, with $10$ initial random samples for training surrogate, and 20 samples acquired in batches of $b=5$.}
    \label{fig:quadruped-baseline}
\end{figure}

\cref{fig:quadruped-baseline} compares our approach against \baselinename{BAMS} and \baselinename{Random} on the \textsf{Quadruped} task, where failures are defined as scenarios $x \in \mathbb{R}^3$ with average velocity tracking error exceeding $0.7$ (e.g., early stopping, large trajectory deviation; see supplementary video). Our method consistently discovers more diverse failures than both baselines (\cref{fig:quadruped-scatter}), and identifies more failures across severity levels for $\gamma = 0.7$ (\cref{tab:quadruped-ps}). In contrast, \baselinename{BAMS} concentrates near the threshold and often misses high-severity failures. We observe that diverse severity coverage also improves surrogate prediction. Our method achieves mean accuracy of $0.80$ over validation failure sets at $\gamma \in \{0.7, 0.8, 0.9, 1.0\}$, compared to $0.70$ for \baselinename{BAMS}, with $0.68$ accuracy on high-severity scenarios ($y_r \geq 1.0$). \cref{fig:quadruped-heatmap} shows that our surrogate captures two symmetric failure modes over $(v_y, w_z)$, consistent with the task structure, confirming that our coverage-aware acquisition actively identifies distinct failures.  

\begin{table}[t]
\centering
\caption{Positive samples for different $\gamma$ values in \textsf{Quadruped} task.}
\label{tab:quadruped-ps}
\begin{tabular}{lcccccc}
\toprule
% & \multicolumn{3}{c}{Our} & \multicolumn{3}{c}{BAMS} \\
% \cmidrule(lr){2-4} \cmidrule(lr){5-7}
$\gamma$
& 0.7 (Our) & 0.8 (Our)& 1.0 (Our)
& 0.7 (BAMS)& 0.8 (BAMS)& 1.0 (BAMS)\\
\midrule
P.S.
& 0.6 $\pm$ 0.1 & 0.3 $\pm$ 0.1 & 0.2 $\pm$ 0.1
& 0.5 $\pm$ 0.1 & 0.3 $\pm$ 0.2 & 0.0 $\pm$ 0.0 \\
\bottomrule
\end{tabular}
\end{table}

\begin{table}[t]
\centering
\setlength{\tabcolsep}{1.5pt}
\renewcommand{\arraystretch}{0.90}
\caption{KITTI component ablation}
% \vspace{-1 mm}
\label{tab:kitti_component_ablation}

\begin{tabular*}{\columnwidth}{@{\extracolsep{\fill}}lccccccc@{}}

\toprule
 & Full & Greedy & $\beta=0 (n+k)$ & $\beta{=}0 $ & MI & $-$MI-C & $-$MI-G \\
\midrule
Utility
& 0.38 & 0.38 & 0.36 & 0.36 & 0.32 & 0.4 & 0.41 \\
P.S.-0.5
& 0.24 & 0.22 & 0.14 & 0.17 & 0.11 & 0.30 & 0.32 \\
P.S.-0.7 
& 0.03 & 0.01 & 0.02 & 0.01 & 0.02 & 0.02 & 0.01 \\
Cov.
& 0.78 & 0.58 & 0.79 & 0.72 & 0.91 & 0.6 & 0.48 \\
\bottomrule
\end{tabular*}

\vspace{-4 mm}

\end{table}

% \begin{table}[t]
% \centering
% \caption{Positive samples for different $\gamma$ values in \textsf{Quadruped} task.}
% \label{tab:quadruped-ps}
% \begin{tabular}{lcccc}
% \toprule
% Method & $\gamma=0.7$ & $\gamma=0.8$ & $\gamma=0.9$ & $\gamma=1.0$  \\
% \midrule
% Our & 0.6 $\pm$ 0.1 & 0.3 $\pm$ 0.1& 0.2 $\pm$ 0.2 & 0.2 $\pm$ 0.1 \\
% BAMS & 0.5 $\pm$ 0.1 & 0.3 $\pm$ 0.2 & 0.1 $\pm$ 0.1  & 0.0 $\pm$ 0.0\\
% \bottomrule
% \end{tabular}
% \end{table}

\subsection{\textsf{SIMPLER}}\label{sec:simpler}

\begin{figure}
    \centering
    \includegraphics[width=0.45\linewidth]{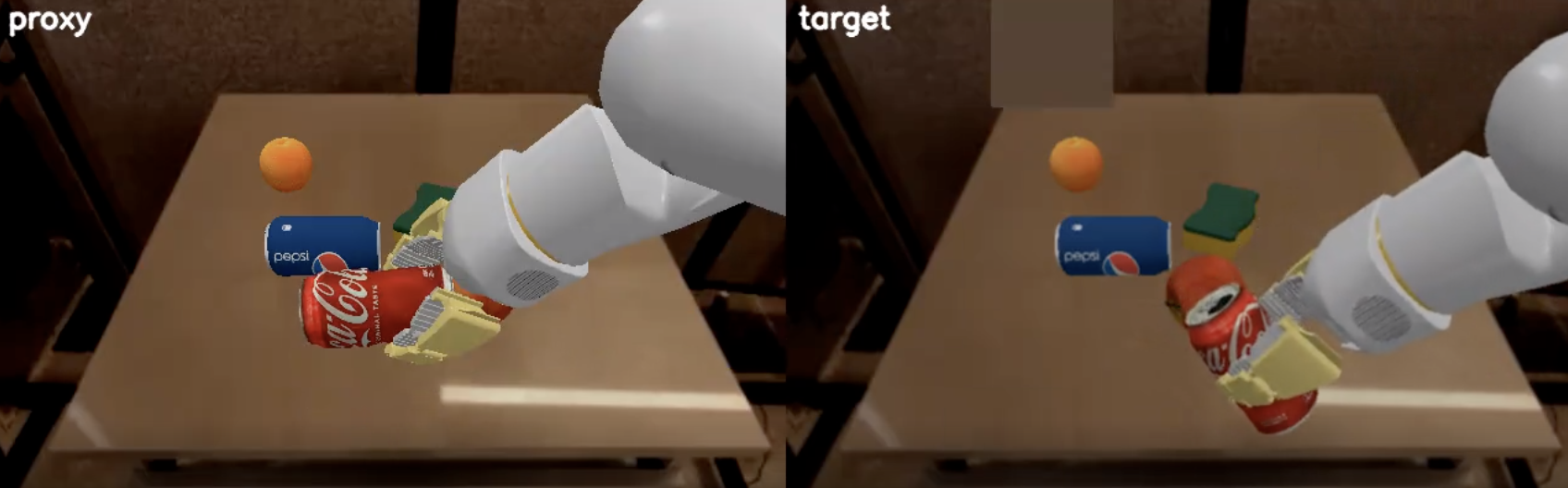}
    \includegraphics[width=0.45\linewidth]{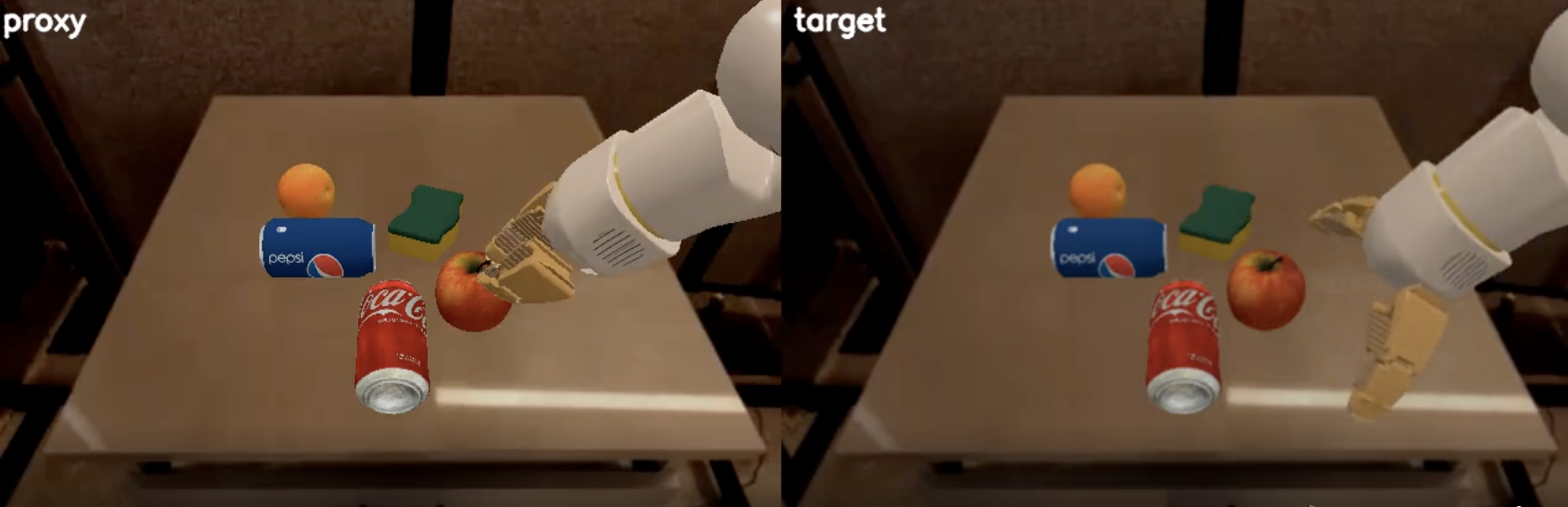}
    \includegraphics[width=0.45\linewidth]
    {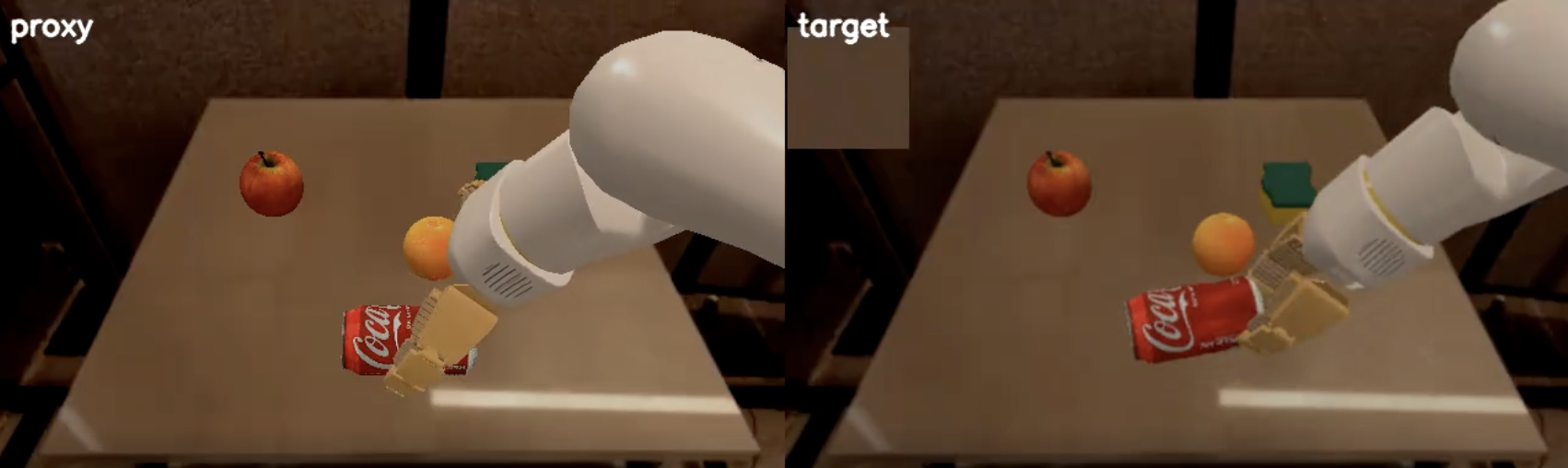}
    \includegraphics[width=0.45\linewidth]{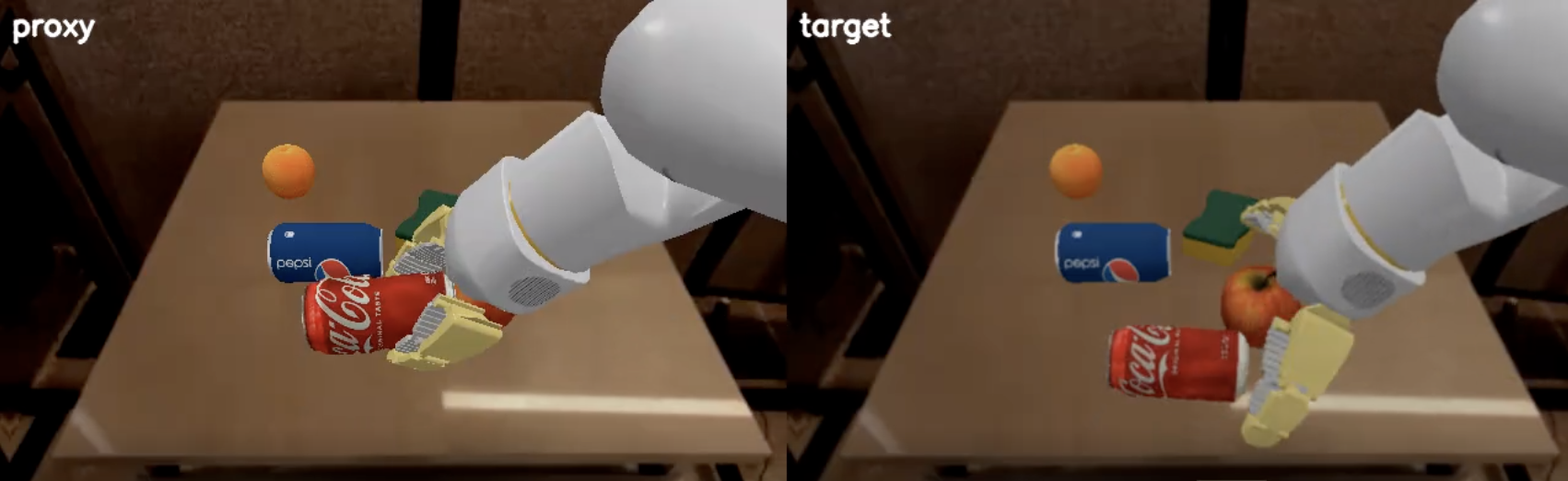}
    
    \caption{\textbf{Examples of grasping failures for \textsf{SIMPLER} task.} Top- scenarios for which both proxy and target systems succeed (left) and fail (right) to grasp the can. Bottom- two examples for which the proxy system succeeds and target system fails to grasp the can, discovered by our approach. }
    \label{fig:simpler}
\end{figure}

% \begin{figure}
%     \centering
%     \includegraphics[width=\linewidth]{figs/quadruped_results.pdf}
%     \caption{\textbf{Baseline comparison} for  \textsf{Quadruped} task, showing avg $\pm$ std across four seeds for $B=30$ samples, with $10$ initial random samples for training all surrogate models, and 20 additional samples acquired in batches of $b=5$.}
%     \label{fig:quadruped-baseline}
% \end{figure}

\cref{fig:simpler-metrics} shows baseline comparison, with \baselinename{BNN-CV} performing comparable to \baselinename{Random} sampling. This is due to the fact that the failure metric of this task is sparsely defined as success ($y_r=0.0$) or failure ($y_s=1.0$), so the lack of exploration leads to heavy dependence on the failures present in initial dataset for accurately learning the surrogate model.
 From the collected failure scenarios, we observe that perceptually challenging conditions with visual clutter are the dominant driver of failure all seeds. We discover two dominant modes for camera positioning to the right and left of reference positions, each corresponding to object placement to the left and far from the robot. The average proxy success rate is $20\%$ showing reasonable sim to real gap. \cref{fig:simpler} (bottom) shows failure scenarios corresponding to this gap with success in proxy system, and failure in target system.

\subsection{KITTI}
We use this task to highlight the merit of our approach with combined sampling against target only sampling, and quantify the contribution of proxy assisted evaluation. Specifically, we test whether proxy information improves acquisition over surrogate-only sampling, using $\beta=0,(n+k)$ (Table~\ref{tab:kitti_component_ablation}), where the surrogate receives all $n+k$ samples, and proxy and $q_{\theta}$ sampling have the same cost; therefore any gains reflect proxy information, not reduced MC error. Importantly, the proxy provides additional system information, not a cheaper substitute for MC samples from $q_{\theta}$. At equal sample count, proxy-assisted acquisition improves failure discovery and utility, preserving coverage, achieving nearly $70\%$ higher coverage of meaningful failure modes, and increases surrogate failure recall rate from $0.18$ to $0.49$. 

We also compare with $\beta=0$ ($n$ surrogate samples). On nuPlan, weak proxy--surrogate correlation in failure regions (avg. $\rho\approx0.15$) drives $\beta\to0$, yielding small gains over no proxy ablation. On KITTI, $\rho\approx0.5$, therefore gains in P.S. and utility are clearer in Table~\ref{tab:kitti_component_ablation}. It can be shown that proxy information can reduce target prediction risk by capturing target variation hidden by $\mathcal{X}$, whereas additional $q_\theta$ samples only reduce variance. This benefit increases with proxy-target correlation. Thus, $\beta$-weighted proxy guidance exposes failure regions missed by surrogate-only sampling and improves the learned target surrogate. 

\section{Discussion \& Conclusion}
Our approach scales gracefully with evaluation budget and scenario complexity, validating our central hypothesis. Baseline comparisons underscore the importance of each component. \baselinename{BAMS} degrades at $170$ evaluations in \textsf{nuPlan} as GP surrogates struggle to scale; a known limitation of GPs noted in \cite{parashar2024failure,sinha2025rate,parashar2025cost}. \baselinename{BNN-CV}'s myopic acquisition collapses onto a narrow failure region in \textsf{SIMPLER}, where the sparse failure metric demands exploration, and \baselinename{GP-C}'s assumption that all proxy failures transfer to the target causes it to miss unpaired failures, despite focusing on diversity of scenarios. Our work shows that proxy corrected risk estimation and scenario diversity are intertwined.  Lack of diversity leads to redundant failures, and without accurate risk estimates, evaluations are wasted. Coupling both components is essential for sample efficient discovery of distinct, severe target-system failures.

 % The consistent performance of our approach across tasks validates our hypothesis, that the method scales gracefully with evaluation budget and scenario complexity. The baseline comparisons underscore the importance of each component. \baselinename{BAMS} performs competitively on the low-dimensional \textsf{Quadruped} task but degrades at the scale of $200$ evaluations in \textsf{nuPlan} due to GP-based surrogates evaluation. \baselinename{BNN-CV}'s myopic, exploitation-only acquisition is most harmful in \textsf{SIMPLER}, where the failure metric is sparse and greedy sampling collapses onto a narrow failure region, leaving large parts of the scenario space unexplored. \baselinename{GP-C} targets diverse sampling but assumes that all proxy failures transfer to the target, missing out on several failures that are not paired. Together, these observations confirm that accurate proxy-corrected failure estimation and scenario diversity are intertwined. Without diversity, even accurate risk estimates yield redundant failures, and without accurate risk estimates, diverse sampling wastes budget on non-critical scenarios. Our results demonstrate that tightly coupling these two components is essential for sample-efficient discovery of distinct, severe target system failures.
\section{Limitations}\label{sec:limitations}

% The primary limitation of our approach is that efficiency remains tied to the quality and dimensionality of the scenario representation, which we assume is pre-defined in our data acquisition. While our method extends failure discovery to higher-dimensional spaces for scalable testing when needed, it does not fundamentally reduce the number of evaluations required as the search space grows. Larger budgets are still needed for high-dimensional scenarios. This also affects surrogate learning as complex scenario representations require more target evaluations to train an accurate surrogate, which may limit applicability in settings where target evaluations are extremely scarce. For such tasks, it is more critical to ensure efficiency of scenario representation, which remains an important direction for future work.

Our approach assumes a scenario representation in which locality, support, and diversity are meaningful. While such representations are available in our experiments, performance may depend on how well they capture behaviorally relevant differences between scenarios. Like other black-box failure discovery methods, our approach also requires target-system evaluations; larger or more complex scenario spaces may require larger budgets to learn accurate target-risk surrogates. Our method is therefore best viewed as a way to use limited target evaluations more effectively by combining proxy guidance, local correction, and diversity-aware acquisition, with future work exploring learned representations and stronger transfer across related systems.

\bibliography{main}

@article{li2026pragmatic,
  title={Pragmatic Curiosity: A Hybrid Learning-Optimization Paradigm via Active Inference},
  author={Li, Yingke and Parashar, Anjali and Zhou, Enlu and Fan, Chuchu},
  journal={arXiv preprint arXiv:2602.06104},
  year={2026}
}

@inproceedings{geiger2012kitti,
  title={Are We Ready for Autonomous Driving? {T}he {KITTI} Vision Benchmark Suite},
  author={Geiger, Andreas and Lenz, Philip and Urtasun, Raquel},
  booktitle={Proceedings of the IEEE Conference on Computer Vision and Pattern Recognition (CVPR)},
  pages={3354--3361},
  year={2012}
}

@article{cabon2020virtual,
  title={Virtual {KITTI} 2},
  author={Cabon, Yohann and Murray, Naila and Humenberger, Martin},
  journal={arXiv preprint arXiv:2001.10773},
  year={2020}
}

@article{anwar2025efficient,
  title={Efficient evaluation of multi-task robot policies with active experiment selection},
  author={Anwar, Abrar and Gupta, Rohan and Merchant, Zain and Ghosh, Sayan and Neiswanger, Willie and Thomason, Jesse},
  journal={arXiv preprint arXiv:2502.09829},
  year={2025}
}

@inproceedings{dawson2023a,
title={A {B}ayesian approach to breaking things: efficiently predicting and repairing failure modes via sampling},
author={Charles Dawson and Chuchu Fan},
booktitle={7th Annual Conference on Robot Learning},
year={2023},
url={https://openreview.net/forum?id=fNLBmtyBiC}
}

@incollection{esposito2005adaptive,
  title={Adaptive {RRT}s for validating hybrid robotic control systems},
  author={Esposito, Joel M and Kim, Jongwoo and Kumar, Vijay},
  booktitle={Algorithmic foundations of robotics vi},
  pages={107--121},
  year={2005},
  publisher={Springer}
}

@inproceedings{dawson2022robust,
  title={Robust counterexample-guided optimization for planning from differentiable temporal logic},
  author={Dawson, Charles and Fan, Chuchu},
  booktitle={IEEE/RSJ International Conference on Intelligent Robots and Systems (IROS)},
  pages={7205--7212},
  year={2022}
}

@inproceedings{ren2023adaptsim,
  title={AdaptSim: Task-Driven Simulation Adaptation for Sim-to-Real Transfer},
  author={Ren, Allen Z and Dai, Hongkai and Burchfiel, Benjamin and Majumdar, Anirudha},
  booktitle={Conference on Robot Learning},
  pages={3434--3452},
  year={2023},
  organization={PMLR}
}

@article{sinha2020neural,
  title={Neural bridge sampling for evaluating safety-critical autonomous systems},
  author={Sinha, Aman and O'Kelly, Matthew and Tedrake, Russ and Duchi, John C},
  journal={Advances in Neural Information Processing Systems},
  volume={33},
  pages={6402--6416},
  year={2020}
}

@inproceedings{dreossi2015efficient,
  title={Efficient guiding strategies for testing of temporal properties of hybrid systems},
  author={Dreossi, Tommaso and Dang, Thao and Donz{\'e}, Alexandre and Kapinski, James and Jin, Xiaoqing and Deshmukh, Jyotirmoy V},
  booktitle={NASA Formal Methods: 7th International Symposium, NFM 2015, Pasadena, CA, USA, April 27-29, 2015, Proceedings 7},
  pages={127--142},
  year={2015},
  organization={Springer}
}

@inproceedings{corso2020scalable,
  title={Scalable autonomous vehicle safety validation through dynamic programming and scene decomposition},
  author={Corso, Anthony and Lee, Ritchie and Kochenderfer, Mykel J},
  booktitle={2020 IEEE 23rd International Conference on Intelligent Transportation Systems (ITSC)},
  pages={1--6},
  year={2020},
  organization={IEEE}
}

@inproceedings{corso2019adaptive,
  title={Adaptive stress testing with reward augmentation for autonomous vehicle validatio},
  author={Corso, Anthony and Du, Peter and Driggs-Campbell, Katherine and Kochenderfer, Mykel J},
  booktitle={2019 IEEE Intelligent Transportation Systems Conference (ITSC)},
  pages={163--168},
  year={2019},
  organization={IEEE}
}

@ARTICLE{10669181,
  author={Parashar, Anjali and Yin, Ji and Dawson, Charles and Tsiotras, Panagiotis and Fan, Chuchu},
  journal={IEEE Robotics and Automation Letters}, 
  title={Learning-based Bayesian Inference for Testing of Autonomous Systems}, 
  year={2024},
  volume={},
  number={},
  pages={1-8},
  doi={10.1109/LRA.2024.3455782}}

@article{rainforth2024modern,
  title={Modern Bayesian experimental design},
  author={Rainforth, Tom and Foster, Adam and Ivanova, Desi R and Bickford Smith, Freddie},
  journal={Statistical Science},
  volume={39},
  number={1},
  pages={100--114},
  year={2024},
  publisher={Institute of Mathematical Statistics}
}

@inproceedings{parashar2024failure,
  title={Failure Prediction from Few Expert Demonstrations},
  author={Parashar, Anjali and Garg, Kunal and Zhang, Joseph and Fan, Chuchu},
  booktitle={NeurIPS 2024 Workshop on Bayesian Decision-making and Uncertainty}
}

@misc{unitree_go2pro,
  author       = {{Unitree Robotics}},
  title        = {{Unitree Go2 Pro Quadruped Robot}},
  year         = {2026},
  howpublished = {\url{https://www.unitree.com/go2}},
  note         = {Quadruped robot platform. Accessed: 2026-05-28}
}

@misc{unitree_rl_mjlab,
  author       = {{Unitree Robotics}},
  title        = {{Unitree RL Mjlab}},
  year         = {2026},
  howpublished = {\url{https://github.com/unitreerobotics/unitree_rl_mjlab}},
  note         = {Reinforcement learning implementation for Unitree robots based on MuJoCo. Accessed: 2026-05-28}
}

@article{parashar2026seed,
  title={SEED-SET: Scalable Evolving Experimental Design for System-level Ethical Testing},
  author={Parashar, Anjali and Li, Yingke and Yu, Eric Yang and Chen, Fei and Neidhoefer, James and Upadhyay, Devesh and Fan, Chuchu},
  journal={arXiv preprint arXiv:2603.01630},
  year={2026}
}

@misc{open_x_embodiment_rt_x_2023,
title={Open {X-E}mbodiment: Robotic Learning Datasets and {RT-X} Models},
author = {Open X-Embodiment Collaboration and Abby O'Neill and Abdul Rehman and Abhinav Gupta and Abhiram Maddukuri and Abhishek Gupta and Abhishek Padalkar and Abraham Lee and Acorn Pooley and Agrim Gupta and Ajay Mandlekar and Ajinkya Jain and Albert Tung and Alex Bewley and Alex Herzog and Alex Irpan and Alexander Khazatsky and Anant Rai and Anchit Gupta and Andrew Wang and Andrey Kolobov and Anikait Singh and Animesh Garg and Aniruddha Kembhavi and Annie Xie and Anthony Brohan and Antonin Raffin and Archit Sharma and Arefeh Yavary and Arhan Jain and Ashwin Balakrishna and Ayzaan Wahid and Ben Burgess-Limerick and Beomjoon Kim and Bernhard Schölkopf and Blake Wulfe and Brian Ichter and Cewu Lu and Charles Xu and Charlotte Le and Chelsea Finn and Chen Wang and Chenfeng Xu and Cheng Chi and Chenguang Huang and Christine Chan and Christopher Agia and Chuer Pan and Chuyuan Fu and Coline Devin and Danfei Xu and Daniel Morton and Danny Driess and Daphne Chen and Deepak Pathak and Dhruv Shah and Dieter Büchler and Dinesh Jayaraman and Dmitry Kalashnikov and Dorsa Sadigh and Edward Johns and Ethan Foster and Fangchen Liu and Federico Ceola and Fei Xia and Feiyu Zhao and Felipe Vieira Frujeri and Freek Stulp and Gaoyue Zhou and Gaurav S. Sukhatme and Gautam Salhotra and Ge Yan and Gilbert Feng and Giulio Schiavi and Glen Berseth and Gregory Kahn and Guangwen Yang and Guanzhi Wang and Hao Su and Hao-Shu Fang and Haochen Shi and Henghui Bao and Heni Ben Amor and Henrik I Christensen and Hiroki Furuta and Homanga Bharadhwaj and Homer Walke and Hongjie Fang and Huy Ha and Igor Mordatch and Ilija Radosavovic and Isabel Leal and Jacky Liang and Jad Abou-Chakra and Jaehyung Kim and Jaimyn Drake and Jan Peters and Jan Schneider and Jasmine Hsu and Jay Vakil and Jeannette Bohg and Jeffrey Bingham and Jeffrey Wu and Jensen Gao and Jiaheng Hu and Jiajun Wu and Jialin Wu and Jiankai Sun and Jianlan Luo and Jiayuan Gu and Jie Tan and Jihoon Oh and Jimmy Wu and Jingpei Lu and Jingyun Yang and Jitendra Malik and João Silvério and Joey Hejna and Jonathan Booher and Jonathan Tompson and Jonathan Yang and Jordi Salvador and Joseph J. Lim and Junhyek Han and Kaiyuan Wang and Kanishka Rao and Karl Pertsch and Karol Hausman and Keegan Go and Keerthana Gopalakrishnan and Ken Goldberg and Kendra Byrne and Kenneth Oslund and Kento Kawaharazuka and Kevin Black and Kevin Lin and Kevin Zhang and Kiana Ehsani and Kiran Lekkala and Kirsty Ellis and Krishan Rana and Krishnan Srinivasan and Kuan Fang and Kunal Pratap Singh and Kuo-Hao Zeng and Kyle Hatch and Kyle Hsu and Laurent Itti and Lawrence Yunliang Chen and Lerrel Pinto and Li Fei-Fei and Liam Tan and Linxi "Jim" Fan and Lionel Ott and Lisa Lee and Luca Weihs and Magnum Chen and Marion Lepert and Marius Memmel and Masayoshi Tomizuka and Masha Itkina and Mateo Guaman Castro and Max Spero and Maximilian Du and Michael Ahn and Michael C. Yip and Mingtong Zhang and Mingyu Ding and Minho Heo and Mohan Kumar Srirama and Mohit Sharma and Moo Jin Kim and Muhammad Zubair Irshad and Naoaki Kanazawa and Nicklas Hansen and Nicolas Heess and Nikhil J Joshi and Niko Suenderhauf and Ning Liu and Norman Di Palo and Nur Muhammad Mahi Shafiullah and Oier Mees and Oliver Kroemer and Osbert Bastani and Pannag R Sanketi and Patrick "Tree" Miller and Patrick Yin and Paul Wohlhart and Peng Xu and Peter David Fagan and Peter Mitrano and Pierre Sermanet and Pieter Abbeel and Priya Sundaresan and Qiuyu Chen and Quan Vuong and Rafael Rafailov and Ran Tian and Ria Doshi and Roberto Mart{'i}n-Mart{'i}n and Rohan Baijal and Rosario Scalise and Rose Hendrix and Roy Lin and Runjia Qian and Ruohan Zhang and Russell Mendonca and Rutav Shah and Ryan Hoque and Ryan Julian and Samuel Bustamante and Sean Kirmani and Sergey Levine and Shan Lin and Sherry Moore and Shikhar Bahl and Shivin Dass and Shubham Sonawani and Shubham Tulsiani and Shuran Song and Sichun Xu and Siddhant Haldar and Siddharth Karamcheti and Simeon Adebola and Simon Guist and Soroush Nasiriany and Stefan Schaal and Stefan Welker and Stephen Tian and Subramanian Ramamoorthy and Sudeep Dasari and Suneel Belkhale and Sungjae Park and Suraj Nair and Suvir Mirchandani and Takayuki Osa and Tanmay Gupta and Tatsuya Harada and Tatsuya Matsushima and Ted Xiao and Thomas Kollar and Tianhe Yu and Tianli Ding and Todor Davchev and Tony Z. Zhao and Travis Armstrong and Trevor Darrell and Trinity Chung and Vidhi Jain and Vikash Kumar and Vincent Vanhoucke and Vitor Guizilini and Wei Zhan and Wenxuan Zhou and Wolfram Burgard and Xi Chen and Xiangyu Chen and Xiaolong Wang and Xinghao Zhu and Xinyang Geng and Xiyuan Liu and Xu Liangwei and Xuanlin Li and Yansong Pang and Yao Lu and Yecheng Jason Ma and Yejin Kim and Yevgen Chebotar and Yifan Zhou and Yifeng Zhu and Yilin Wu and Ying Xu and Yixuan Wang and Yonatan Bisk and Yongqiang Dou and Yoonyoung Cho and Youngwoon Lee and Yuchen Cui and Yue Cao and Yueh-Hua Wu and Yujin Tang and Yuke Zhu and Yunchu Zhang and Yunfan Jiang and Yunshuang Li and Yunzhu Li and Yusuke Iwasawa and Yutaka Matsuo and Zehan Ma and Zhuo Xu and Zichen Jeff Cui and Zichen Zhang and Zipeng Fu and Zipeng Lin},
howpublished  = {\url{https://arxiv.org/abs/2310.08864}},
year = {2023},
}

@article{li24simpler,
          title={Evaluating Real-World Robot Manipulation Policies in Simulation},
          author={Xuanlin Li and Kyle Hsu and Jiayuan Gu and Karl Pertsch and Oier Mees and Homer Rich Walke and Chuyuan Fu and Ishikaa Lunawat and Isabel Sieh and Sean Kirmani and Sergey Levine and Jiajun Wu and Chelsea Finn and Hao Su and Quan Vuong and Ted Xiao},
          journal = {arXiv preprint arXiv:2405.05941},
          year={2024},
          }

@inproceedings{karnchanachari2024towards,
  title={Towards learning-based planning: The nuplan benchmark for real-world autonomous driving},
  author={Karnchanachari, Napat and Geromichalos, Dimitris and Tan, Kok Seang and Li, Nanxiang and Eriksen, Christopher and Yaghoubi, Shakiba and Mehdipour, Noushin and Bernasconi, Gianmarco and Fong, Whye Kit and Guo, Yiluan and others},
  booktitle={2024 IEEE International Conference on Robotics and Automation (ICRA)},
  pages={629--636},
  year={2024},
  organization={IEEE}
}

@article{cao2021bayesian,
  title={Bayesian active learning by disagreements: A geometric perspective},
  author={Cao, Xiaofeng and Tsang, Ivor W},
  journal={arXiv preprint arXiv:2105.02543},
  year={2021}
}

@article{Chaloner1995BayesianReview,
    title = {{Bayesian experimental design: A review}},
    year = {1995},
    journal = {Statistical science},
    author = {Chaloner, Kathryn and Verdinelli, Isabella},
    pages = {273--304}
}

@inproceedings{parashar2025cost,
  title={Cost-aware Discovery of Contextual Failures using Bayesian Active Learning},
  author={Parashar, Anjali and Zhang, Joseph and Li, Yingke and Fan, Chuchu},
  booktitle={9th Annual Conference on Robot Learning},
  year={2025}
}

@article{badithela2025reliable,
  title={Reliable and scalable robot policy evaluation with imperfect simulators},
  author={Badithela, Apurva and Snyder, David and Zha, Lihan and Mikhail, Joseph and O'Kelly, Matthew and Dixit, Anushri and Majumdar, Anirudha},
  journal={arXiv preprint arXiv:2510.04354},
  year={2025}
}

@inproceedings{luo2025leveraging,
  title={Leveraging correlation across test platforms for variance-reduced metric estimation},
  author={Luo, Rachel and Yang, Heng and Watson, Michael and Sharma, Apoorva and Veer, Sushant and Schmerling, Edward and Pavone, Marco},
  booktitle={9th Annual Conference on Robot Learning}
}

@inproceedings{sinha2025rate,
  title={Rate-Informed Discovery via Bayesian Adaptive Multifidelity Sampling},
  author={Sinha, Aman and Nikdel, Payam and Paul, Supratik and Whiteson, Shimon},
  booktitle={Conference on Robot Learning},
  pages={2579--2598},
  year={2025},
  organization={PMLR}
}

\newpage

\appendix
\tableofcontents

 \begin{figure}[t]
    \centering
    \includegraphics[width=\linewidth]{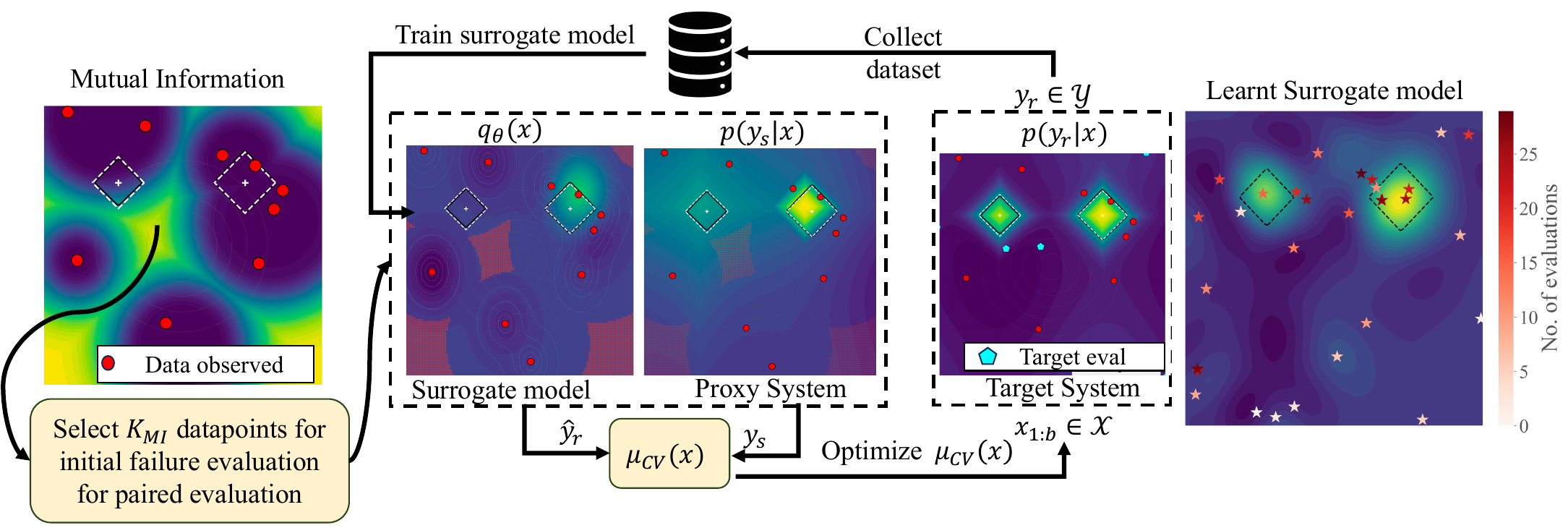}
    \caption{\textbf{Overview of approach.} Our method uses support-aware Mutual Information (\cref{eq:MI}) to subsample $K_{MI}$ scenarios $x$, which are evaluated on the cheap proxy system to get metrics $y_s$, and a surrogate model $q_{\theta}$ trained using data evaluated on target system. We subsequently optimize $\mu_{CV}(x)$ (\cref{eq:mu_cv})over these sub sampled candidates to select the next batch of evaluation points as likely failure scenarios. The process is repeated until evaluation budget $B$ is exhausted. Figure on the left shows the surrogate model learnt after 40 samples of evaluation for the target system shown.}
    \label{fig:process-2d}
\end{figure}
\section{Local minimum-variance property of the control-variate estimator}\label[appendix]{app:cv-global}

We show that the estimator in \cref{eq:mu_cv} is the minimum-variance estimator within the class of local affine control-variate estimators. Specifically, our method does not reduce surrogate bias, but uses proxy system information in a meaningful way to accelerate efficiency of failure discovery. Our proof follows proof of Theorem 1 from \citep{luo2025leveraging}. Fix a candidate scenario $x\in\mathcal{X}$, and let $B_x\subseteq\mathcal{X}$ denote the local neighborhood used to estimate the target failure statistic around $x$. Let
 $Y_r\sim p(y_r\mid x')$, $Y_s\sim p(y_s\mid x')$, and $x'\sim B_x$.  In practice, the target samples $Y_r$ are replaced by surrogate predictions $\hat{Y}_r\sim q_\theta(x')$, so the following result is interpreted conditional on the learned surrogate $q_\theta$.

Suppose we have $n$ paired local samples
\begin{equation}
    \{(\hat{y}_r^i,y_s^i)\}_{i=1}^n,
\end{equation}
and $k$ additional proxy-only samples
\begin{equation}
    \{y_s^j\}_{j=n+1}^{n+k}.
\end{equation}
Consider the class of local affine control-variate estimators
\begin{equation}
    \mu_{\mathrm{CV}}(\beta)
    =
    \mu_{CV}(x) = \frac{1}{n} \sum_{i=1}^{n} (\hat{y}_r^i-\boldsymbol{\beta} y_s^i )
+ \frac{1}{k} \sum_{j=1}^{k} \beta y_s^i 
    \label{eq:local-cv-class}
\end{equation}
Assuming that the paired and proxy-only samples are drawn independently from the same local neighborhood distribution over $B_x$, this estimator is unbiased for the local target statistic:
\begin{align}
    \mathbb{E}_{B_x}
    \left[
    \hat{\mu}_{\mathrm{CV}}(\beta)
    \right]
    &=
    \mathbb{E}_{B_x}
    \left[
    Y_r-\beta^\top Y_s
    \right]
    +
    \mathbb{E}_{B_x}
    \left[
    \beta^\top Y_s
    \right] \\
    &=
    \mathbb{E}_{B_x}[Y_r].
\end{align}
Thus, the choice of $\beta$ affects the variance but not the expectation.

For a fixed $\beta$, independence of the paired and proxy-only sample sets gives
\begin{equation}
    \mathrm{Var}
    \left(
    \hat{\mu}_{\mathrm{CV}}(\beta)
    \right)
    =
    \frac{1}{n}
    \mathrm{Var}
    \left(
    Y_r-\beta^\top Y_s
    \right)
    +
    \frac{1}{k}
    \mathrm{Var}
    \left(
    \beta^\top Y_s
    \right).
    \label{eq:cv-var-start}
\end{equation}
Expanding the two variance terms, we obtain
\begin{align}
    \mathrm{Var}
    \left(
    \mu_{\mathrm{CV}}(\beta)
    \right)
    &=
    \frac{1}{n}\mathrm{Var}(Y_r)
    -
    \frac{2}{n}
    \beta^\top \mathrm{Cov}(Y_s,Y_r)
    +
    \frac{1}{n}
    \beta^\top \mathrm{Var}(Y_s)\beta
    +
    \frac{1}{k}
    \beta^\top \mathrm{Var}(Y_s)\beta \\
    &=
    \frac{1}{n}\mathrm{Var}(Y_r)
    -
    \frac{2}{n}
    \beta^\top \mathrm{Cov}(Y_s,Y_r)
    +
    \frac{n+k}{nk}
    \beta^\top \mathrm{Var}(Y_s)\beta .
    \label{eq:cv-var-expanded}
\end{align}
This is a convex quadratic function of $\beta$ whenever $\mathrm{Var}(Y_s)$ is positive semidefinite, and is strictly convex when $\mathrm{Var}(Y_s)$ is positive definite. Differentiating \cref{eq:cv-var-expanded} with respect to $\beta$ and setting the derivative equal to zero gives
\begin{equation}
    -\frac{2}{n}\mathrm{Cov}(Y_s,Y_r)
    +
    \frac{2(n+k)}{nk}
    \mathrm{Var}(Y_s)\beta
    =
    0.
\end{equation}
Therefore, the variance-minimizing coefficient is
\begin{equation}
    \beta^\star
    =
    \frac{k}{n+k}
    \mathrm{Var}(Y_s)^{-1}
    \mathrm{Cov}(Y_s,Y_r).
    \label{eq:beta-star-local}
\end{equation}
Substituting $\beta^\star$ into \cref{eq:local-cv-class} gives the local control-variate estimator used in \cref{eq:cv-estimator}. Hence, $\mu_{\mathrm{CV}}$ is the minimum-variance estimator among all estimators of the affine local control-variate form in \cref{eq:local-cv-class}.

The corresponding optimal variance is
\begin{equation}
    \mathrm{Var}
    \left(
    \mu_{\mathrm{CV}}(\beta^\star)
    \right)
    =
    \frac{1}{n}
    \left(
    \mathrm{Var}(Y_r)
    -
    \frac{k}{n+k}
    \mathrm{Cov}(Y_r,Y_s)
    \mathrm{Var}(Y_s)^{-1}
    \mathrm{Cov}(Y_r,Y_s)
    \right).
    \label{eq:local-cv-opt-var}
\end{equation}
In the scalar case, this can be written as
\begin{equation}
    \mathrm{Var}
    \left(
    \mu_{\mathrm{CV}}(\beta^\star)
    \right)
    =
    \frac{\mathrm{Var}(F)}{n}
    \left(
    1
    -
    \frac{k}{n+k}
    \rho^2(Y_r,Y_s)
    \right),
    \label{eq:local-cv-rho}
\end{equation}
where $\rho(Y_r,Y_s)$ is the local Pearson correlation between the target and proxy statistics. Since $\rho^2(Y_r,Y_s)\in[0,1]$, we have
\begin{equation}
    \mathrm{Var}
    \left(
    \hat{\mu}_{\mathrm{CV}}(\beta^\star)
    \right)
    \leq
    \frac{\mathrm{Var}(Y_r)}{n},
\end{equation}
which is the variance of the direct local Monte Carlo estimator using only the $n$ surrogate induced-statistic samples over $Y_r$. Therefore, whenever the proxy statistic is locally correlated with the surrogate induced statistic, the control-variate estimator strictly reduces variance. When the local correlation is zero, it matches the target-only estimator in variance.

Applying this result to our setting, $Y_r$ corresponds to the local target failure statistic induced by the surrogate $q_\theta$, and $Y_s$ corresponds to the local proxy statistic induced by $p(y_s\mid x)$. Thus, under the local sampling assumptions above and conditional on the learned surrogate, the coefficient in \cref{eq:beta-star-local} yields the minimum-variance local affine estimator of the target failure statistic. This justifies using $\mu_{\mathrm{CV}}$ as a proxy-corrected failure predictor for selecting scenarios likely to satisfy $\mu_r(x)\geq \gamma$. 

\paragraph{Note:} For MC sampled data $x'\sim B_x$, the resulting $\hat{y}_r \sim q_{\theta}(x)$ may give biased results in practice as the data used to train $q_{\theta}(x)$ is not unbiased, and acquired through active learning, hence this justification is only valid for the case when surrogate induced statistic is an unbiased representation of the target statistic. To bridge this practical gap, we focus on selecting right hyperparameter for the training and architecture of $q_{\theta}$. Newer training paradigms that calibrate the loss of training 
$q_{\theta}$, such that the resulting estimator is equivalent to an estimator trained on unbiased data will be investigated as a part of our future work.

\section{Motivation for MI-adjusted data acquisition}\label[appendix]{app:MI-efe}
We combine MI with failure discovery as two distinct objectives in data acquisition. This setup emerges naturally under active inference based data acquisition proposed by \citep{li2026pragmatic}. In \cref{app:MI-hyperparam}, we discuss the construction of various terms that appear in \cref{sec:mi} in detail.

\section{Defining locality in the feature space}\label[appendix]{app:local-region}

This section discusses the details of construction of local neighborhood region $\mathcal{B}_x$ in detail for the case where $\mathcal{X}$ is a continuous space.  

\begin{figure}
     \centering
     \includegraphics[width=\linewidth]{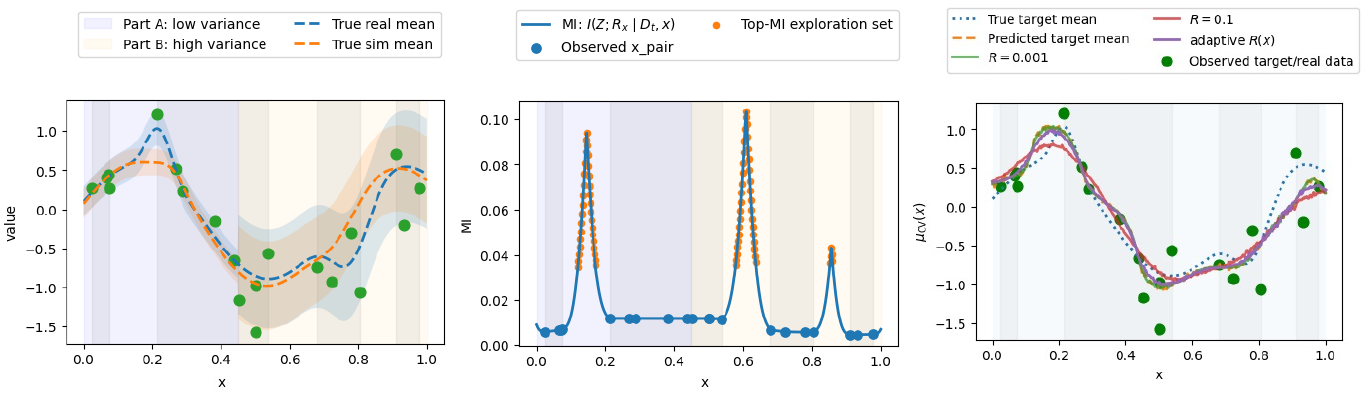}
     \caption{\textbf{1D synthetic task.} Left-right: (1) Randomly sampled 10 scenarios and observed values of synthetic target, shown against ground truth target and proxy system. (2) Scenarios identified as high MI due to being outside support/novel candidates (\cref{sec:mi}). (3) $\hat{\mu}_{CV}(x)$ estimated using different fixed local regions and adaptive local region selection plotted against ground truth target. Adaptively selecting local region shows minimum deviation from ground truth.}
     \label{fig:cv-radius}
 \end{figure}

% The motivation behind using $\mathcal{B}_x$ to perform localized sampling is driven by the construction of conditional mean estimates in Prediction  Powered Inference (PPI), which is a special case of Control variate architecture with $\beta=1$. The use of neighborhood data to construct a local mean estimator is especially useful when training data is limited, impacting  the accuracy of $q(x)$. 
The neighborhood $\mathcal{B}_x$ can be defined in multiple ways, and controls the granularity of $\mu_{CV}(x)$. 
For example, consider a continuous region, with the neighborhood $\mathcal{B}_x$  defined using a Euclidean ball with radius $R$ as $\mathcal{B}_x = \{x'|\lVert x-x'\rVert^2_2 \leq R\}$. Consider, $R \in [R_{\text{min}},R_{\text{max}}]$, such that for $R=R_{\text{max}}$, $\mathcal{B}_x \equiv \mathcal{X}$. The corresponding $\mu_{CV}(x)$ is equivalent to a global average CV estimator, represented in \cite{luo2025leveraging}. On the other hand, prediction for $R \to 0$ collapses to a point-wise local mean prediction. 
% For a surrogate model trained with sufficient data, local point-wise prediction accurately represents point-wise mean, but is not the case for a surrogate model trained with insufficient data. 
For cases where we cannot manually specify the region $\mathcal{B}_x$, we propose a sensitivity aware heuristic for adaptively adjusting $R \in [R_{\text{min}},R_{\text{max}}]$, such that if the local sensitivity of prediction is high, $R$ should be smaller and vice versa. We define a sensitivity term $s(x)$ that measures sensitivity of prediction using local gradient information and variance of the surrogate model at $x$:
\begin{equation}
s(x)
=
\frac{
\left\|\nabla_x \mu_{q}(x)\right\|
}{
\left(\sigma_{q}(x) + \epsilon\right)
}, \quad \quad 
\tilde{s}(x)
=
\frac{s(x) - \min_{x' \in \mathcal{X}} s(x')}
{\max_{x' \in \mathcal{X}} s(x') - \min_{x' \in \mathcal{X}} s(x') + \epsilon}.
\end{equation}
Here $\tilde{s}(x) \in [0,1]$ is a normalized version of $s(x)$ for practical application, and $\sigma_q^2(x) = \text{Var}_{q_{\theta}(x)}[y_r]$.  We construct our heuristic as:

\begin{equation}\label{eq:r-a}
    R_a(x)
=
R_{\min}
+
\frac{\left(R_{\max} - R_{\min}\right)}{1 + \exp\left(\tilde{s}(x)\right)}.
\end{equation}
\cref{fig:cv-radius} shows $\mu_{CV}$ as a function of $x$ for a 1D toy example, with local region $\mathcal{B}_x$ chosen using fixed radius values and adaptive $ R_a(x)$ chosen using \cref{eq:r-a}. We observe that the prediction with $R_a$ most accurately predicts the ground-truth target mean as opposed to fixed region alternatives.

% Incorporating $\mu_{CV}$ in our data acquisition strategy enables prediction of likely failure scenarios by efficiently combining both sources of evaluation. An additional data acquisition goals is to prioritize scenario novelty for target system evaluation. We view it as optimizing for scenarios that would significantly update belief about existing training support $\mathcal{X}_t = \{x^i\}_{i=1}^t$. Below we propose an approach that uses clustering to estimate training support, which can be used for discovery of novel scenarios via MI. 

\section{Construction of terms for MI estimation}\label[appendix]{app:MI-hyperparam}

To estimate MI, we need to estimate entropy $H$, for which, we model the distribution over $R_x$ as $p(R_x|x,\mathcal{D}_t) = \sum_{k=1}^Kp(R_x|Z_x=k) p(Z_x=k|x,\mathcal{D}_t)$. Here, prior $p(Z_x=k|\mathcal{D}_t,x)$ controls probability of assignment of a point $x$ to a cluster, and can be either observed directly for probabilistic clustering approaches such as GMMs, or can be manually designed, as shown below for K-means clustering used in this work. We define the likelihood $p(R_x|Z_x=k)$ to be high for scenarios far from existing clusters as:
\begin{equation}\label{eq:p_R}
    p(R_x=1 \mid Z_x=k)=q_k(x),
\qquad
p(R_x=1 \mid Z_x=\mathrm{new}, x, D_t)=q_{\mathrm{new}}(x),
\end{equation}
Here $ 0\leq q_k(x) \leq \varepsilon_{\mathrm{exist}}$ is a distance weighted distribution that is $0$ close to an existing cluster, and is capped at a small value $\varepsilon_{\mathrm{exist}}$ far away. On the other hand, $q_{\mathrm{new}}(x)$ is defined as:
\begin{equation}\label{eq:p_R_q_new}
    q_{\mathrm{new}}(x)
=
\sigma\left(
\frac{d_{\min}(x)-\rho}{\gamma}
\right),
\qquad
\sigma(z)=\frac{1}{1+e^{-z}},
\end{equation}
where $d_{\min}(x)$ is the distance from $x$ to the nearest observed cluster boundary, $\gamma$ is a softness parameter and $\rho$ is a novelty radius which controls the minimum distance threshold for a scenario $x$ to be considered novel. \cref{eq:p_R} and \cref{eq:p_R_q_new} specify a model for geometric distance based support discovery such that $p(R_x=1)$ has high probability mass around novel scenarios, leading to the discovery of new clusters. We use data adaptive approaches in practice to estimate $\gamma,\rho$ and $K$ to mitigate hyperparameter dependency. $K$ is estimated from a range of $[K_{\text{min}},K_{\text{max}}]$ using the elbow heuristic. We derive
$\gamma,\rho$ directly from the geometry of the current training support
$\mathcal{D}_t$. Specifically, we compute the nearest-neighbour
distance for each support point $x_i \in \mathcal{D}_t$ as

\begin{equation}
    d_{\mathrm{nn},i} = \min_{j \neq i} \|x_i - x_j\|,
\end{equation}
and set
\begin{equation}
    \rho = \mathrm{median}\!\left(\{d_{\mathrm{nn},i}\}_{i=1}^{M}\right),
    \qquad
    \gamma = \mathrm{std}\!\left(\{d_{\mathrm{nn},i}\}_{i=1}^{M}\right),
\end{equation}

\subsection*{Defining probability $p(Z_x\mid x,\mathcal{D}_t)$}
Given cluster centers $\{c_k\}_{k=1}^{K}$ with per-cluster RMS spread
$l_k = \sqrt{\frac{1}{|k|}\sum_{i \in k} \|x_i - c_k\|^2}$, the prior
over support state $Z_x$ at  $x$ is defined using a distance based kernel $w_k$:

\begin{equation}
    w_k(x) = \exp\!\left( -\frac{\|x - c_k\|^2}{2\,l_k^2} \right),
    \qquad
    W(x) = \sum_{k=1}^{K} w_k(x)
\end{equation}

\begin{equation}
    p(Z = z_k \mid x) = \frac{w_k(x)}{W(x) + 1},
    \qquad
    p(Z = z_{\mathrm{new}} \mid x) = \frac{1}{W(x) + 1}
\end{equation}

\cref{fig:process-2d} shows the reconstruction of failure regions using a GP surrogate model by performing data acquisition using our approach on a tor 2D synthetic task. The task consists of two low prob failures shown in yellow, which we aim to be discover in $B=40$ samples. \cref{fig:process-2d} shows that our overall approach succesfully applies to uncover both regions, showing that our MI strategy helps in searching for diverse failures, and $\mu_{\text{CV}}$ helps to bridge sim and real gap.

\section{nuPlan ablations}\label[appendix]{app:nuplan-ablations}
% \begin{figure}
%     \centering
%     \includegraphics[width=\linewidth]we a
%     \caption{\textbf{Visualization of dominant failure modes}.\textsf{nuPlan}. T-SNE plots of scenarios selected by each method plotted against top-four dominant modes. Mode boundaries visualize convex hulls for scenarios with closed loop TTC $\leq 0.5$. }
%     \label{fig:nuplan-visualize}
% \end{figure}
\subsection{Ablation results for size of initial dataset}

We conduct ablation for initial size of sampling with $N_0=5,10,20,50$ initial data points for surrogate model training for our method, results for $B=100$ samples reported in \cref{tab:nuplan-ablation-1}. Clearly, for $N_0=50$ we observe lowest cumulative TTC and highest diversity as well as failure count, however, we observe that the difference between $N_0=5,10,20$ is marginal, which tells us that in the lack of sufficient initial data, our method's performance is not substantially affected. For another 20 samples, with $B=140$, the run initialized with $N_0=20$ reports $1.05, 0.8$ and $0.33$ for TTC average, coverage and positive samples, indicating that the lack of sufficient initial samples in high dimensional scenarios can be compensated for by increasing the testing budget and collecting more sequential data. 

\begin{table}[t]
\centering
\caption{Ablation over different initialization sizes on the \textsf{nuPlan} task, 100 samples, four seeds, avg $\pm$ std reported.}
\label{tab:nuplan-ablation-1}
\begin{tabular}{lccc}
\toprule
Setting & Avg. Cumulative TTC & Cumulative Coverage & Positive Samples \\
\midrule
$n_{\text{init}}=5$  & $1.09 \pm 0.06$ & $0.58 \pm 0.04$ & $0.27 \pm 0.02$ \\
$n_{\text{init}}=10$ & $1.16 \pm 0.13$ & $0.48 \pm 0.09$ & $0.25 \pm 0.01$ \\
$n_{\text{init}}=20$ & $1.06 \pm 0.04$ & $0.75 \pm 0.02$ & $0.26 \pm 0.03$ \\
$n_{\text{init}}=50$ & $0.99 \pm 0.06$ & $1.10 \pm 0.13$ & $0.30 \pm 0.05$ \\
\bottomrule
\end{tabular}
\end{table}

\subsection{Ablation results for $\mu_{\text{CV}}$}
\cref{tab:nuplan-ablation-beta} shows results for ablation over \cref{eq:mu_cv} with $\beta=0$, to remove the dependency on sim samples. We observe that with $\beta=0$, there is marginal improvement in positive samples and cumulative TTC, however, this comes at the cost of coverage, which is especially compromised for samples with TTC $=0.0$, i.e., severe failure scenarios. This is due to the fact that while proxy scenarios, in this case constructed from open loop simulation, are a noisy estimate of the target system, having larger sampling capacity on proxy helps identify more regions where the target system might be failing, which may be missed in limited budget target only estimation. 

\begin{table}[t]
\centering
\caption{Ablation comparing our approach with and without sim samples by setting $\beta=0$ on the \textsf{nuPlan} task. Results reported for 20 initial samples, $B=150$, avg and std for four seeds.}
\label{tab:nuplan-ablation-beta}
\begin{tabular}{lccc}
\toprule
Setting & Avg. Cumulative TTC & Cumulative Coverage & Positive Samples \\
\midrule
\multicolumn{4}{l}{$\gamma=0.3$} \\
\midrule
nominal & $1.04 \pm 0.0332$ & $0.98 \pm 0.06$ & $0.32 \pm 0.01$ \\
$\beta=0$  & $1.03 \pm 0.05$ & $0.95 \pm 0.05$ & $0.35 \pm 0.01$ \\
\midrule
\multicolumn{4}{l}{$\gamma=0.0$} \\
\midrule
nominal & $1.04 \pm 0.03$ & $0.87 \pm 0.05$ & $0.29 \pm 0.01$ \\
$\beta=0$  & $1.03 \pm 0.05$ & $0.81 \pm 0.08$ & $0.31 \pm 0.01$ \\
\bottomrule
\end{tabular}
\end{table}
\section{Quadruped}\label[appendix]{app:quadruped}

\begin{figure}
    \centering
    \includegraphics[width=\linewidth]{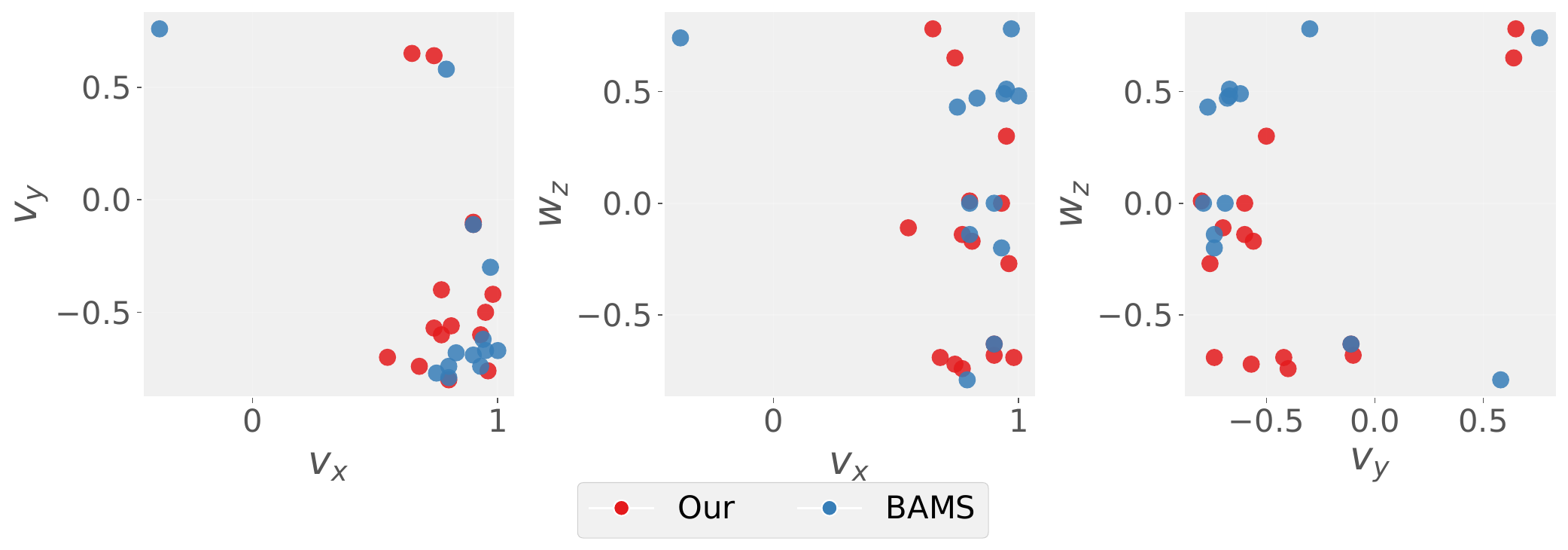}
    \caption{Scatterplot of failure scenarios $x = (v_x,v_y,w_z)$ ($y\geq 0.7$) for \textsf{Quadruped} task, generated by our method (red), and \baselinename{BAMS} (blue). Our method shows higher diversity of scenarios unlike \baselinename{BAMS} that tends to concentrate around failures discovered in early iterations.}
    \label{fig:quadruped-scatter}
\end{figure}

\begin{figure}
    \centering
    \includegraphics[width=\linewidth]{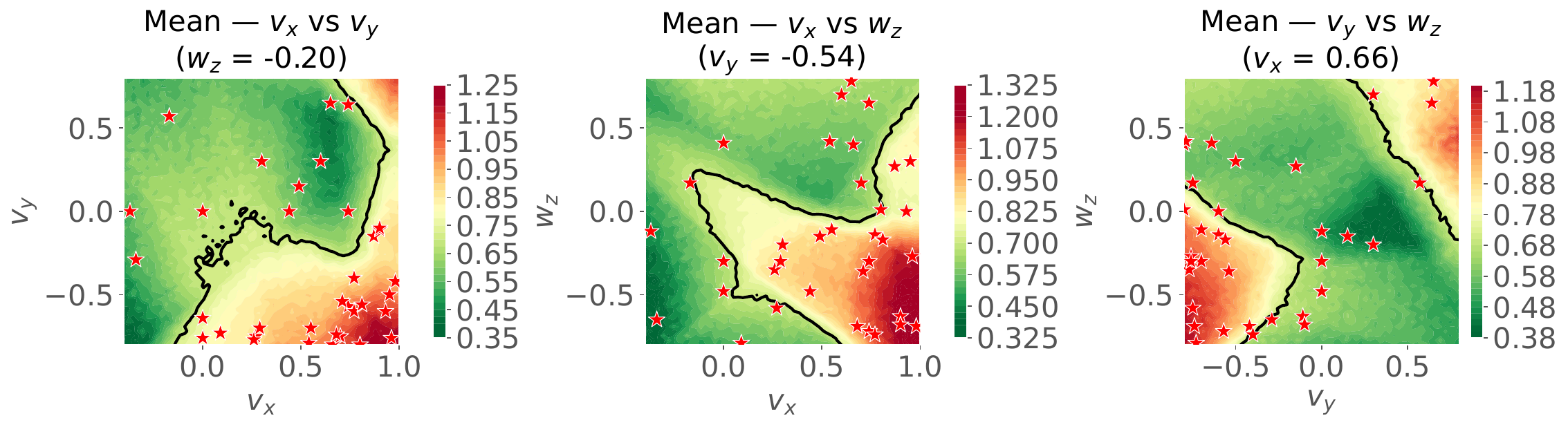}
    \caption{Heatmaps showing learnt surrogate model predictions projected over scenario space for \textsf{Quadruped} task. Projection value corresponds to median of third variable reported in parentheses. Scattered points correspond to initial 10 and subsequent 20 scenarios collected by our approach.}
    \label{fig:quadruped-heatmap}
\end{figure}

Additional results for \textsf{Quadruped}, showing scatterplot of our method and \baselinename{BAMS} in \cref{fig:quadruped-scatter}, and \cref{fig:quadruped-heatmap} showing heatmap of learnt failure versus success region, providing qualitative evidence that our method discovers more diverse failures. We also provide video demos of two failure and one non-fail command velocity tracking tasks in the supplementary video attached, where the scenarios are selected as those predicted fail and non fail by our method. 

\section{SIMPLER}
\begin{figure}
    \centering
    \includegraphics[width=\linewidth]{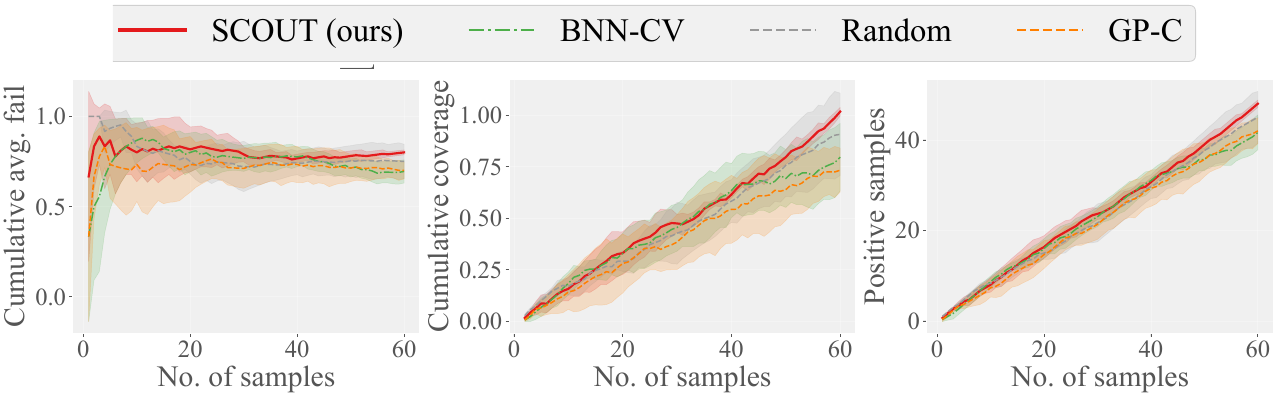}
    \caption{\textbf{Baseline comparison} for \textsf{SIMPLER} task, showing avg $\pm$ std across three seeds for $B=70$ samples, with $10$ initial random samples for training all surrogate models, and 60 additional samples acquired in batches of $b=3$.}
    \label{fig:simpler-metrics}
\end{figure}

\label[appendix]{app:simpler}
Baseline comparison for \textsf{SIMPLER} shown in \cref{fig:simpler-metrics}. We also show videos of fail and non fail tasks, where the successful task is provided as a reference, and fail tasks are sampled by our approach consisting of one task where both proxy and target system fail, and one where only target system fails, showing that our approach can discover failures unseen in simulation as well. 
\section{Visualization}
\begin{figure}
    \centering
    \includegraphics[width=0.3\linewidth]{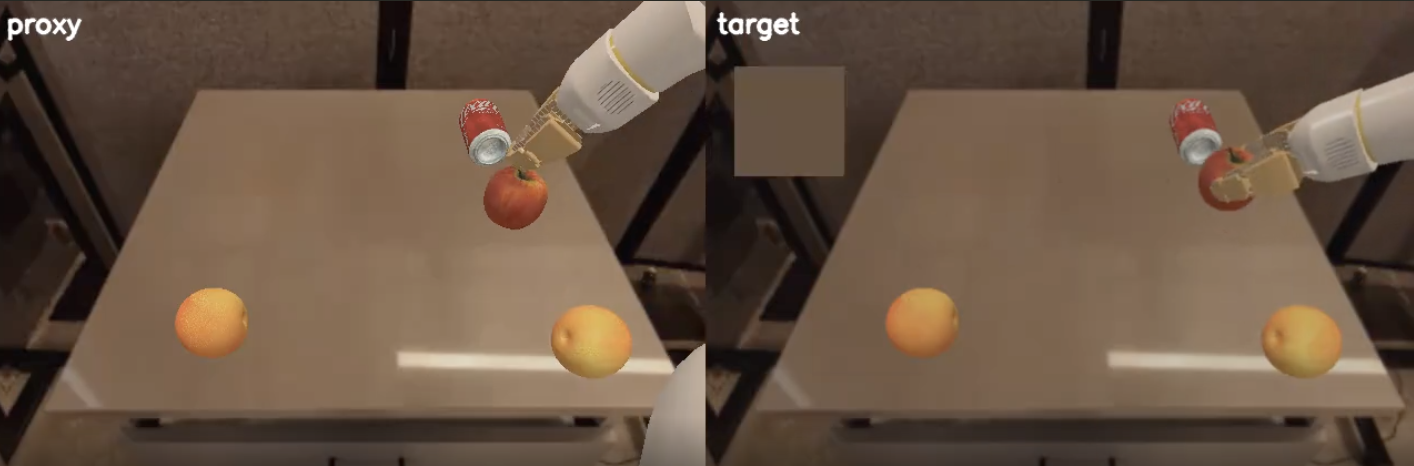}
\includegraphics[width=0.3\linewidth]{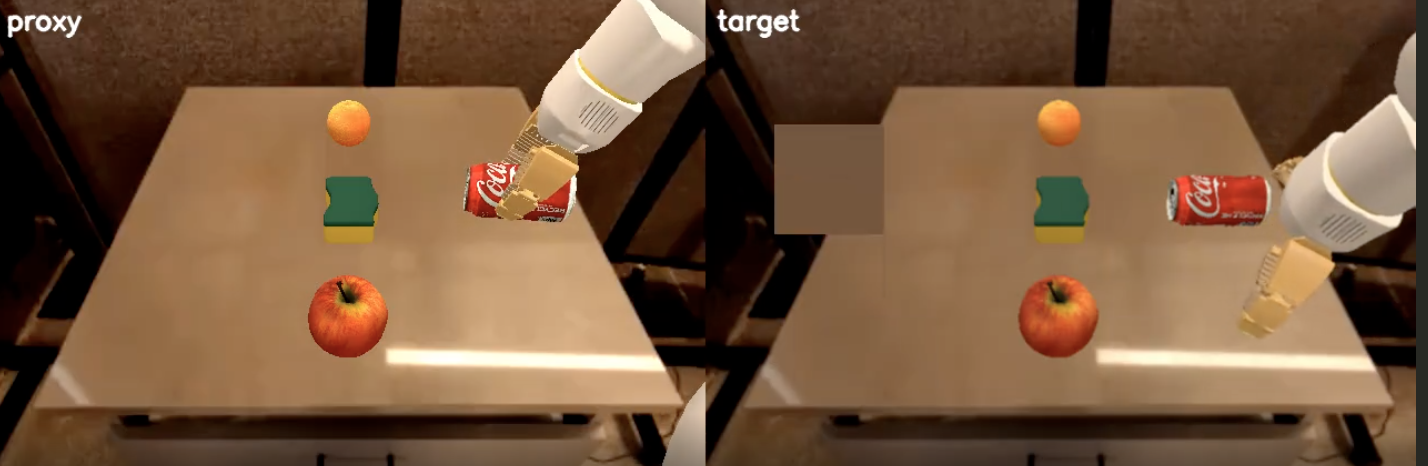}
\includegraphics[width=0.3\linewidth]{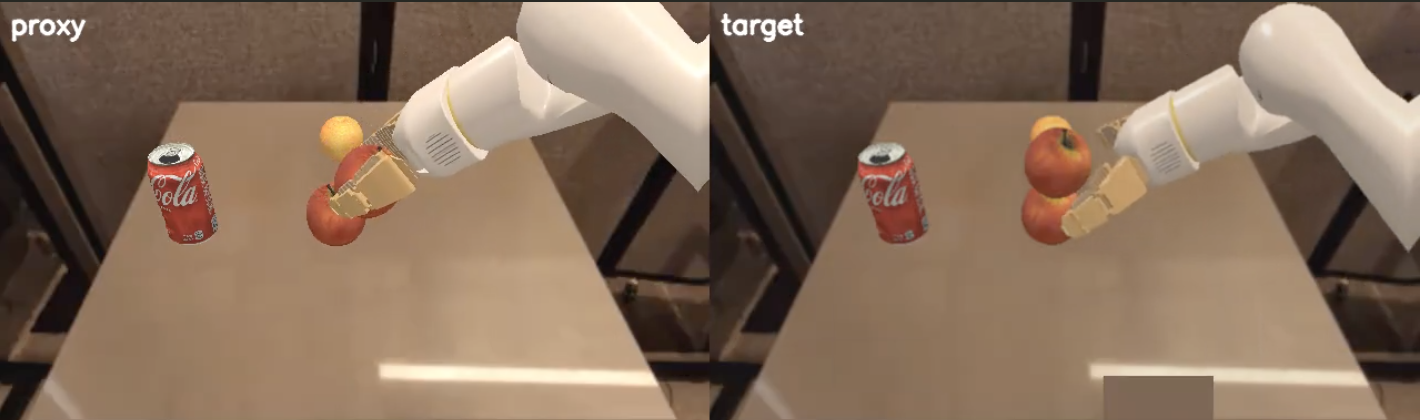}
    \caption{More explains of failure scenarios uncovered by our approach in the \textsf{SIMPLER} task.}
    \label{fig:more-simpler}
\end{figure}

\label[appendix]{app:scenario}
In \cref{fig:more-simpler} we show additional examples for \textsf{SIMPLER} task. We also show videos of failures for \textsf{Quadruped} task, which correspond to high velocity in $v_x$ and $v_y$, as tracking larger requires crossing the obstacles used to construct a boundary, leading to obstacle avoidance getting activated and velocity becoming zero.
\begin{figure}
    \centering
    \includegraphics[width=0.45\linewidth]{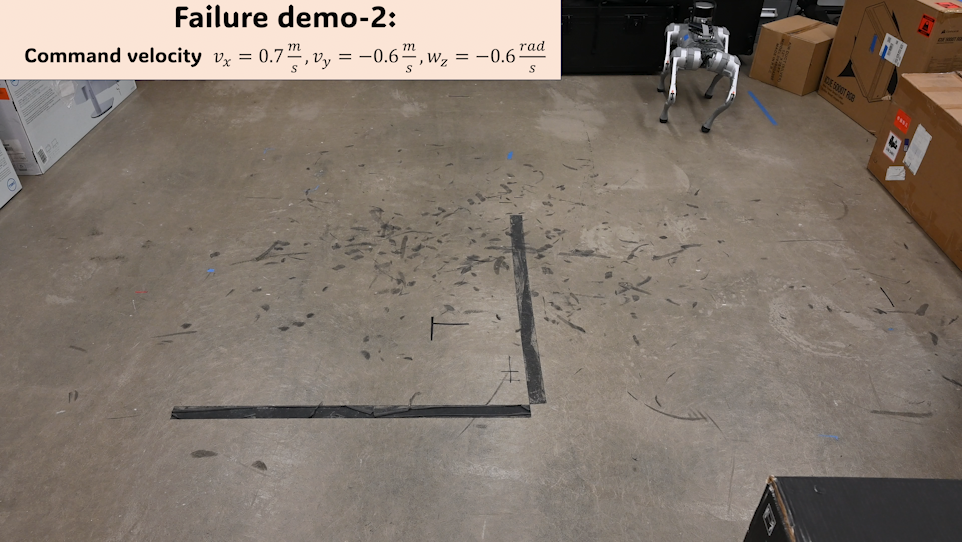}
    \includegraphics[width=0.45\linewidth]{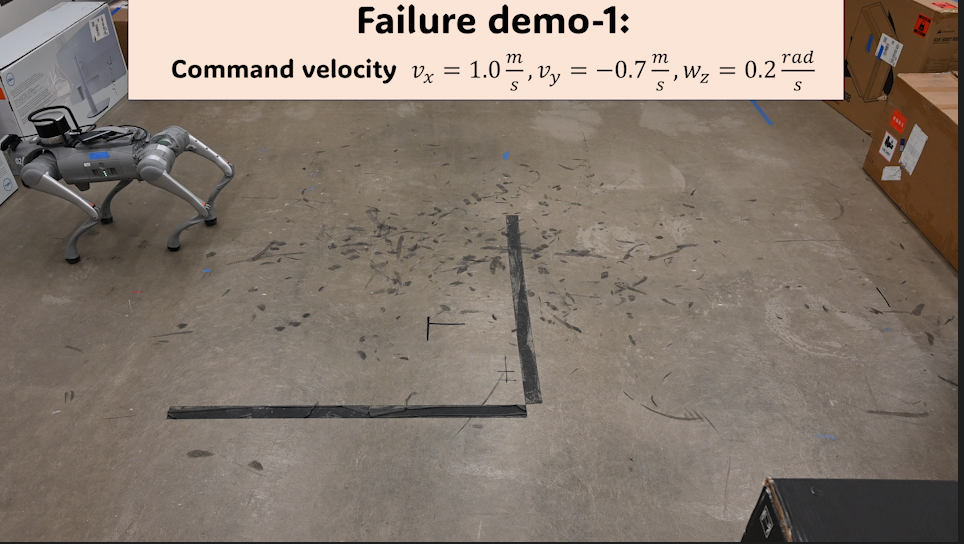}
    \caption{Failures for \textsf{Quadruped} task. Quadrupeds shown at the end of tracking duration, failure scenarios correspond to \textsf{Quadruped} reaching near one of the boxed and stopping due to obstacle avoidance.}
    \label{fig:quadruped-failure}
\end{figure}

\section{Experiment details}\label[appendix]{app:experiment-details}

\subsection{Model specifications}
We choose randomly sampled data to initialize the various surrogate models for each setup. For \textsf{nuPlan} and \textsf{SIMPLER}, we use BNN surrogate models for our method, which consist of a MLP with MC dropouts at each layer ($p(\text{dropout})=0.1$), and for \textsf{Quadruped}, we use MLP, which is suitable choice due to low observed stochasticity in the velocity tracking policy. For each setup and across all baselines that use a NN based architecture, we train using \textsf{Adam} optimizer with lr=1e-3, with 1000, 800 and 3000 epochs for \textsf{nuPlan, SIMPLER} and \textsf{Quadruped} tasks respectively. These hyperparameter are selected based on training accuracy over the initial random sampling dataset. \cref{tab:architectures} summarizes the model architectures for all tasks. \textsf{nuPlan} has inherent over-parameterization in scenario due to encoding so we benefit from a compressive architecture whereas \textsf{SIMPLER} benefits from over-parameterization.

\begin{table}[h]
\centering
\caption{Summary of neural network architectures across tasks.}
\label{tab:architectures}
\begin{tabular}{lllll}
\toprule
Task & Architecture & Input & Hidden layers
    & Activation \\
\midrule
\textsf{nuPlan} surrogate
    & MLP & 384 & $96 \to 24 \to 6$ 
    & ELU + LN  \\
\textsf{nuPlan} BNN-C head
    & + sigmoid & --- & --- 
    & $\sigma$ \\
\textsf{SIMPLER} BNN
    & MLP & 19 & $96 \to 24 \to 6$ 
    & Tanh \\
\textsf{SIMPLER} classifier for GP-C
    & MLP + sigmoid & 19 & $96 \to 24 \to 6$ 
    & $\sigma$  \\
\textsf{Quadruped} MLP
    & MLP & 3 & $8 \to 8$ 
    & ReLU \\
\bottomrule
\end{tabular}
\end{table}

\subsection{Data specifications}
For \textsf{nuPlan}, proxy dataset is  directly available and used as it is, whereas \textsf{SIMPLER} and \textsf{Quadruped} are setups where querying proxy data can be costly as well. Hence, we use a BNN and MLP for constructing a surrogate model for proxy systems for \textsf{SIMPLER} and \textsf{Quadruped}, using larger initial datasets of 50 and 2000 datapoints collected offline. For \textsf{SIMPLER}, the initial dataset for proxy surrogate model is quite less, hence, at each step of acquisition for target system, we also update proxy system with 1 datapoint. \baselinename{BAMS} expects a cost of collecting proxy and target data as an input, which is calibrated to match the frequency of data collection used by our approach.  For \baselinename{BNN-C} in \textsf{nuPlan}, we use the same architecture as ours. Note that the baseline is originally supposed to work with GPs, which can be incompatible for large size of \textsf{nuPlan}, hence we use a BNN there.

\subsection{Scenario design and failure metrics for each task}
For \textsf{nuPlan}, scenarios are driving logs from nuPlan database, which are encoded into 384-dimensional vectors, we measure TTC as failure metric generated by nuPlan, and proxy and target system correspond to open loop and closed loop simulations respectively. 

For \textsf{SIMPLER} task, we perturb visual specifications and object placements, generating a 19 dimensional scenario specification. The target object (a Coke can) is
placed at a 2D tabletop position $x_{\text{obj}} \in [-0.5, -0.1]$\,m
and $y_{\text{obj}} \in [0.0, 0.4]$\,m, spanning the robot's reachable
workspace. Camera viewpoint variation is introduced via lateral
translations $\Delta x_{\text{cam}},\, \Delta y_{\text{cam}} \in
[-0.025, 0.025]$\,m relative to the nominal camera pose. Scene
appearance is controlled by brightness $b \in [0.70, 1.10]$ and
contrast $c \in [0.90, 1.25]$, modelling illumination variation between
the proxy and target domains. Finally, distractor objects are placed at
positions $x_{\text{dist}} \in [-0.5, -0.1]$\,m,
$y_{\text{dist}} \in [0.0, 0.4]$\,m with uniformly sampled yaw
$\theta_{\text{dist}} \in [-\pi, \pi]$\,rad, introducing clutter that
occludes the target object and disrupts grasping. Across three task
variants, \texttt{pick\_horizontal}, \texttt{pick\_vertical}, and
\texttt{pick\_standing}, and choosing distractor objects from a selection of 8 objects makes a 19 dimensional scenario representation.

Failure here is recorded as failure to grasp or lift the coke can, and is generated by the environment. Target and proxy systems correspond to slight difference in visual specifications, and target system has extra visual noise acting as occlusion, leading to worse performance on several scenarios.

For \textsf{Quadruped}, the scenario is 3-d and corresponds to command velocity $v_x,v_y,w_z$ in the range $[-0.8,1.0]\times[-0.8,0.8]\times[-0.8,0.8]$ and proxy system consists of a simulation with RL policy for command velocity tracking, whereas target system consists of the actual hardware with Unitree sports mode used for command velocity tracking. We initialize the quadruped at the same location within a square boundary constructed using cardboard boxes. However, we observe that failure scenarios are agnostic to initialization provided they are initialized at one of the four corners.

\end{document}